\documentclass[runningheads]{llncs}

\usepackage{eccv}

\usepackage{eccvabbrv}
\usepackage{amsmath}
\usepackage{amssymb}
\usepackage{booktabs}
\usepackage{pifont}
\usepackage{algorithm}
\usepackage{algpseudocode}
\usepackage{marvosym}
\usepackage{multirow}
\usepackage{stfloats}
\usepackage{caption}
\usepackage{soul}
\usepackage{colortbl}

\usepackage{graphicx}
\usepackage{booktabs}
\usepackage{wrapfig}

\usepackage[pagebackref, breaklinks, colorlinks, citecolor = eccvblue]{hyperref}
\usepackage{marvosym}

\usepackage[accsupp]{axessibility}  

\usepackage{orcidlink}

\begin{document}

\title{OneRestore: A Universal Restoration Framework for Composite Degradation} 


\author{Yu Guo\inst{1,2,\dagger} \orcidlink{0000-0002-0642-7684} \and %
Yuan Gao\inst{1,\dagger} \orcidlink{0000-0001-7730-0421} \and %
Yuxu Lu\inst{3} \orcidlink{0000-0001-9845-7516} \and %
Huilin Zhu\inst{1,2} \orcidlink{0000-0003-1607-7283} \and %
Ryan Wen Liu\inst{1}$^{(\textrm{\Letter})}$ \orcidlink{0000-0002-1591-5583} \and %
Shengfeng He\inst{2}$^{(\textrm{\Letter})}$ \orcidlink{0000-0002-3802-4644}}%

\authorrunning{Y. Guo et al.}

\institute{Wuhan University of Technology, Wuhan, China \\
\email{wenliu@whut.edu.cn} \and
Singapore Management University, Singapore \\
\email{shengfenghe@smu.edu.sg} \and
The Hong Kong Polytechnic University, Hong Kong, China \\
$^\dagger$ Equal Contribution \\
\url{https://github.com/gy65896/OneRestore}}

\maketitle
\begin{abstract}
    In real-world scenarios, image impairments often manifest as composite degradations, presenting a complex interplay of elements such as low light, haze, rain, and snow. Despite this reality, existing restoration methods typically target isolated degradation types, thereby falling short in environments where multiple degrading factors coexist. To bridge this gap, our study proposes a versatile imaging model that consolidates four physical corruption paradigms to accurately represent complex, composite degradation scenarios. In this context, we propose OneRestore, a novel transformer-based framework designed for adaptive, controllable scene restoration. The proposed framework leverages a unique cross-attention mechanism, merging degraded scene descriptors with image features, allowing for nuanced restoration. Our model allows versatile input scene descriptors, ranging from manual text embeddings to automatic extractions based on visual attributes. Our methodology is further enhanced through a composite degradation restoration loss, using extra degraded images as negative samples to fortify model constraints. Comparative results on synthetic and real-world datasets demonstrate OneRestore as a superior solution, significantly advancing the state-of-the-art in addressing complex, composite degradations.
\end{abstract}

\section{Introduction}
\begin{figure*}[t]
        \centering
        \includegraphics[width=1\linewidth]{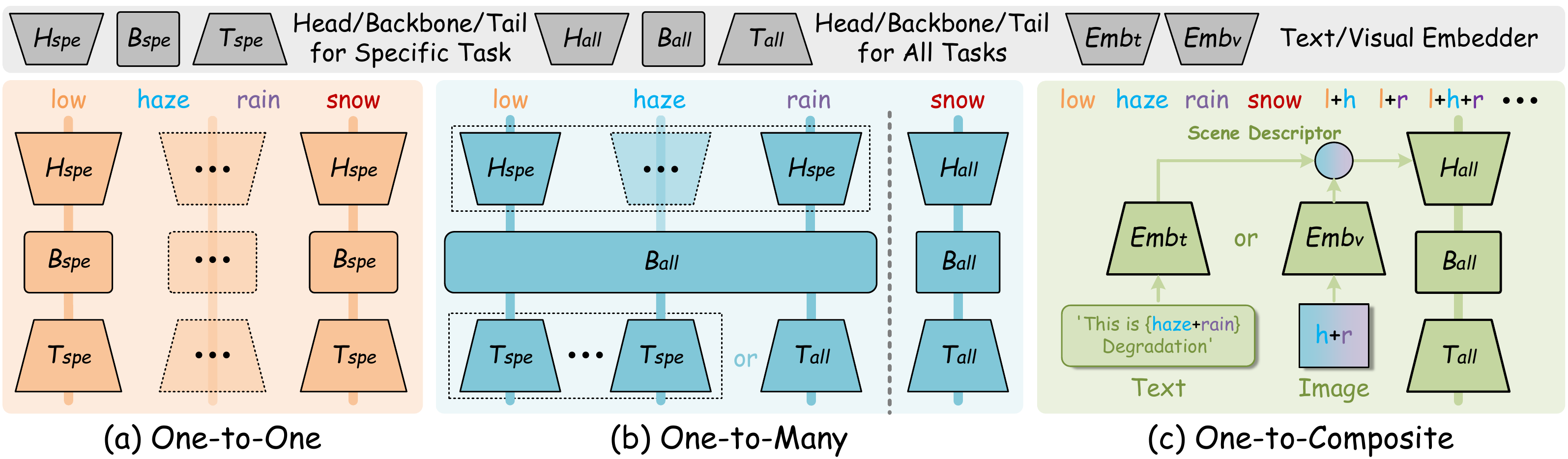}\\
        \vspace{-2mm}
        \caption{Structural comparison between One-to-One, One-to-Many, and our One-to-Composite image restoration methods.}
        \vspace{-4mm}
        \label{fig:compare}
    \end{figure*}

    Image restoration is an essential task that entails the recovery of high-quality visuals from compromised inputs. The fidelity of these restored images is vital for the subsequent downstream tasks, such as robot navigation and autonomous driving. Considerable progress has been made in addressing single degradation scenarios by a single model (\ie, One-to-One models, see Fig.~\ref{fig:compare}a), such as low-light conditions~\cite{zhou2023low, xu2023low, guo2023low}, haze~\cite{song2023vision, hoang2023transer, zheng2023curricular}, rain~\cite{luo2023local, fu2023continual, du2023dsdnet}, and snow~\cite{quan2023image, chen2023lightweight, chen2023msp}. These methods have achieved remarkable success within their intended contexts. Nevertheless, real-world conditions often involve unpredictable and variable composite degradations that can severely impact image quality and clarity~\cite{li2020all}. The occurrence of mixed degrading conditions in one image can lead to intricate interactions, posing significant challenges to One-to-One solutions \cite{jiao2020formnet, zhou2022pro, zamir2021multi, zamir2022restormer}. The necessity for seamless switching and decision-making across specialized methods for each condition proves to be inefficient and ineffective in practical applications. Therefore, there is a growing imperative for robust, all-encompassing image restoration approaches that can universally address multiple forms of degradation through a singular, dynamic framework.
    \begin{figure}[t]
        \centering
        \includegraphics[width=1\linewidth]{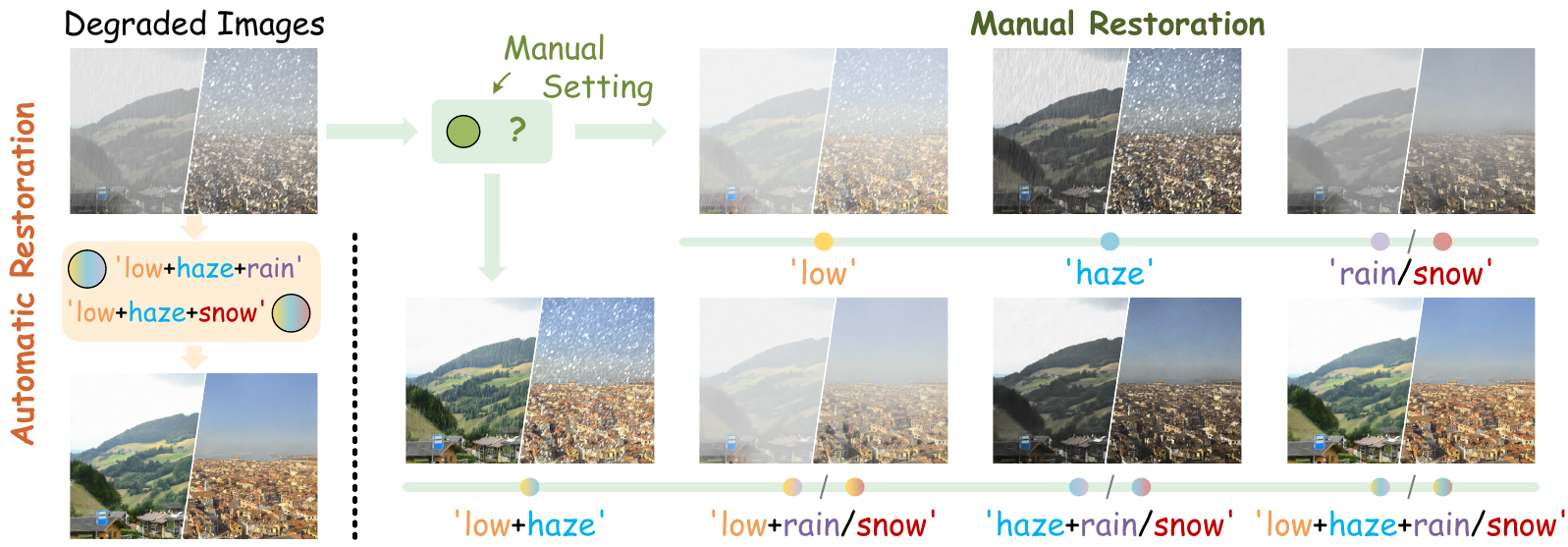}
        \vspace{-6mm}
        \caption{Our OneRestore allows fully controllable image restoration using scene descriptors derived from both automatic visual attribute extraction (top) and manual text embeddings (bottom).}
        \vspace{-5mm}
        \label{fig:performance}
    \end{figure}

    Recent research has been directed toward One-to-Many models, which address various individual degradation factors within a single framework. These models are categorized into: those with partial parameter-sharing (One-to-Many$_p$) \cite{li2020all, chen2021pre, han2022blind, wang2023smartassign} and those with full parameter-sharing (One-to-Many$_f$)~\cite{chen2022learning, wang2023context, valanarasu2022transweather, li2022all, ozdenizci2023restoring}. One-to-Many$_p$ (left of Fig.~\ref{fig:compare}b) restoration methods typically employ models with either multiple heads and tails or a single tail to address various types of image impairments. However, the complexity and size of these models tend to scale with the diversity of degradations they aim to correct, leading to increased training costs and parameter counts. Furthermore, when images suffer from a combination of degrading factors, these models may underperform. Some One-to-Many$_p$ techniques also depend on user intervention to select appropriate restoration weights, which is impractical for automated or dynamic settings. On the other hand, One-to-Many$_f$ (right of Fig.~\ref{fig:compare}b) approaches train models using a direct learning strategy, where inputs are images with different individual degradations and the targets are their clear counterparts. However, this direct training approach is designed predominantly for single-category degradation issues, it does not adequately address the complex dynamics that arise when multiple types of impairments affect an image simultaneously. As a result, such methods can falter in practical applications where user control and precise directionality of restoration are crucial. For instance, in cases where users seek to improve low-light imagery, the model might inadvertently prioritize dehazing actions, deviating from user intent.

    Drawing from the preceding discussion, we delineate the primary challenges in developing image restoration models for composite degradations in real-world conditions:
    \begin{itemize}
    \item  \textit{How can the model be designed to adaptively discern the specific degradation afflicting the input image?}
    \item  \textit{What approaches enable a model to be trained efficiently across a spectrum of degradations while enhancing user control over the restoration process?}
    \item  \textit{How can the recovery process be optimized that not only approach the quality of clear images but also maintain distinction from various degraded forms?}
    \end{itemize}
    In response to the challenges outlined, we present a unified model that builds upon established physical corruption models for degradations such as low light, haze, rain, and snow. This model effectively captures the complexity of real-world degradation conditions and forms the basis for constructing our Composite Degradation Dataset (CDD-11), which serves both training and evaluation purposes. To address the issue of composite degradation, our work introduces \textit{OneRestore}, a transformer-based universal restoration architecture. This novel framework distinguishes itself from conventional methods by embodying a One-to-Composite strategy, as depicted in Fig.~\ref{fig:compare}c, which allows for precise and controllable outcomes (see Fig.~\ref{fig:performance}). It employs an innovative cross-attention block that fuses scene descriptors, which can be manual text embeddings or derived automatically from visual attributes. We further enhance the model's robustness by integrating a composite degradation restoration loss, using additional degraded images as negative samples to establish stringent lower-bound constraints. 
    Our key contributions are summarized as follows:
    \begin{itemize}
        \item We present the first attempt to unify composite degradations, creating a versatile imaging model that simulates a range of degradation types. This innovation underpins the development of our comprehensive Composite Degradation Dataset.
        \item We propose a universal framework that introduces a novel level of controllability in image restoration, employing a cross-attention mechanism that integrates scene descriptors derived from either manual text embeddings or automated visual attribute extractions.
        \item We tailor a composite degradation restoration loss, enhancing the model's ability to discern between various forms of image degradation.
        \item Our universal model not only sets a new state-of-the-art for composite dataset performance but also competes favorably with One-to-One models in single-degradation scenarios.
    \end{itemize}
\section{Related Work}
\label{sec:relat}
    \textbf{Complex Degradation Image Restoration.}
    Recent advancements in image restoration have introduced a variety of solutions for complexly degraded images. A notable direction is the partial parameter-sharing One-to-Many approach, employing architectures with multiple heads and tails, and a single backbone, to address various degradation factors. For instance, Li \textit{et al.}~\cite{li2020all} develop an all-weather network that restores scenes affected by haze, rain, and snow, using specialized encoders for each condition and a shared decoder. Chen \textit{et al.}~\cite{chen2021pre} introduce a transformer-based model for deraining, denoising, and super-resolving images, leveraging a multi-head and multi-tail design. Wang \textit{et al.}~\cite{wang2023smartassign} present a knowledge assignment strategy to differentiate and process rain and snow degradations adaptively. While these models represent strides in complex degradation image restoration, they stop short of being truly universal One-to-Many networks. With only partial weight sharing, they address different degradation types through distinct network pathways.
    
    Conversely, universal image restoration has emerged with the One-to-Many methods sharing all network weights, thereby uniformly mitigating multiple degradation types. Examples include the unified two-stage knowledge learning framework by Chen \textit{et al.}~\cite{chen2022learning}, which efficiently processes dehazing, deraining, and desnowing with a singular pretrained weight set. Li \textit{et al.}~\cite{li2022all} present a network capable of comprehensive image restoration without the need for pre-identifying degradation levels. Additionally, {\"O}zdenizci \textit{et al.}~\cite{ozdenizci2023restoring} introduce a conditional denoising diffusion model tailored for patch-based image restoration under varied weather conditions. However, due to the significant differences and mutual interference between various degradation factors, this type of method can only solve specific types of degradation and cannot achieve image restoration in composite degradation scenarios. To this end, we propose a universal One-to-Composite image restoration method that allows the model to adapt to arbitrary combinations of composite degradations.

    \textbf{Visual Attribute Control.}
    Visual attributes are pivotal in defining image features and enhancing the interpretability of models. They enrich feature representations and provide a hierarchical framework for image analysis. Early research highlighted visual attributes' transformative impact on image generation, recognition, and retrieval~\cite{NIPS2007_ed265bc9, kumar2011describable, yan2016attribute2image}. More recent studies underscore their utility in improving object recognition and detection, particularly in data-scarce environments~\cite{NEURIPS2020_acaa23f7, meng2020adinet, NEURIPS2020_fa2431bf, zhu2020attribute}. Visual attributes have also been leveraged for zero-shot learning applications, advancing classification tasks beyond the constraints of labeled datasets~\cite{saini2022disentangling, wang2023learning}. In our study, we harness visual attributes to delineate the intricate degradation patterns present in real-world imaging conditions, enhancing the pertinence and efficacy of image restoration techniques. It is noteworthy that conventional discriminators, when tasked with identifying image degradation, are typically confined to image data, precluding autonomous feature control. Our approach, by contrast, accommodates both images and text descriptions as inputs, thereby facilitating the automatic or manual determination of degradation characteristics.
\section{Composite Degradation Formulation}
\label{sec:data}
    \subsection{Imaging Formulation}
    In real-world scenarios, images are often affected by multiple degrading factors simultaneously, resulting in suboptimal performance from restoration methods designed for isolated degradations. To address the broad spectrum of degradation scenarios, we introduce a versatile imaging model that encapsulates all degradations, including low light, haze, rain, and snow. This model aims to narrow the disparity between simulated and authentic real-world data. Given a degraded image $I(x)$ and its corresponding clear state $J(x)$, the proposed imaging model is mathematically articulated as
    \begin{equation}
    \label{eq:data}
        I(x) = \mathcal{P}_h(\mathcal{P}_{rs}(\mathcal{P}_l(J(x)))),
    \end{equation}
    where $\mathcal{P}_l$, $\mathcal{P}_{rs}$, and $\mathcal{P}_h$ denote the low light, rain/snow, and haze degradations, respectively. Given that concurrent rain and snow conditions are infrequent, they are not addressed within the scope of this work. A comprehensive exposition of the proposed imaging model is detailed below.
    \textbf{Low-Light Conditions.}
    Inspired by Retinex theory, low-light degraded image $I_l(x)$ can be generated by
    \begin{equation}
    \label{eq:low}
        I_l(x) = \mathcal{P}_l(J(x)) = \frac{J(x)}{L(x)}{L(x)}^\gamma + \varepsilon.
    \end{equation}
    Here, we use a darkening coefficient $\gamma \in [2, 3]$ as the brightness controller to adjust the illumination map $L(x)$ generated via LIME~\cite{guo2016lime}. To simulate poor light conditions realistically, we add the Gaussian noise $\varepsilon$ with mean $= 0$ and variance $\in [0.03, 0.08]$.
    \textbf{Rain/Snow Streaks.} 
    Following the methodologies outlined in~\cite{li2019heavy, chen2021all}, our process superimposes rain or snow streaks onto the image before synthesizing haze. To introduce rain streaks $\mathcal{R}$, we produce the rain-degraded image $I_{rs}(x)$ by overlaying $\mathcal{P}_{rs}$ onto the low-light image $I_l(x)$, which is expressed as
    \begin{equation}
    \label{eq:rain}
        I_{rs}(x) = \mathcal{P}_{rs}(I_l(x)) = I_l(x) + \mathcal{R}.
    \end{equation}
    For snow streaks $\mathcal{S}$, the snow-degraded image is defined as
    \begin{equation}
    \label{eq:snow}
        I_{rs}(x) = \mathcal{P}_{rs}(I_l(x)) = I_l(x)(1-\mathcal{S}) + M(x)\mathcal{S},
    \end{equation}
    with $M(x)$ being the chromatic aberration map. 

    \textbf{Haze Degradations.}
    We incorporate haze degradation by employing the atmospheric scattering model as
    \begin{equation}
    \label{eq:haze}
        I(x) = \mathcal{P}_{h}(I_{rs}(x)) = I_{rs}(x)t+A(1-t),
    \end{equation}
    where $t$ denotes the transmission map and $A$ represents the atmospheric light. The transmission map $t$ is defined by the exponential decay of light, $t=e^{-\beta d(x)}$, where $\beta$ represents the haze density coefficient and $d(x)$ is the scene depth (estimated by MegaDepth~\cite{li2018megadepth}). The haze density coefficient $\beta$ is varied within the range of $[1.0, 2.0]$ and the atmospheric light $A$ is constrained to $[0.6, 0.9]$ for realistic simulation.
 
    \subsection{Composite Degradation Dataset} 
    Drawing on Eq.~\eqref{eq:data}, we have developed the Composite Degradation Dataset (CDD-11), encompassing 11 categories of image degradations and their clear counterparts. These degraded samples include \textit{low} (low-light), \textit{haze}, \textit{rain}, \textit{snow}, \textit{low+haze}, \textit{low+rain}, \textit{low+snow}, \textit{haze+rain}, \textit{haze+snow}, \textit{low+haze+rain}, and \textit{low+haze+snow}. From the RAISE database~\cite{dang2015raise}, we selected 1,383 high-resolution clear images for producing 11 composite degradations. They are resized to a uniform resolution of 1080$\times$720. The overall dataset is split into 13,013 image pairs for training and 2,200 for testing. 
    \begin{figure*}[t]
        \centering
        \includegraphics[width=\linewidth]{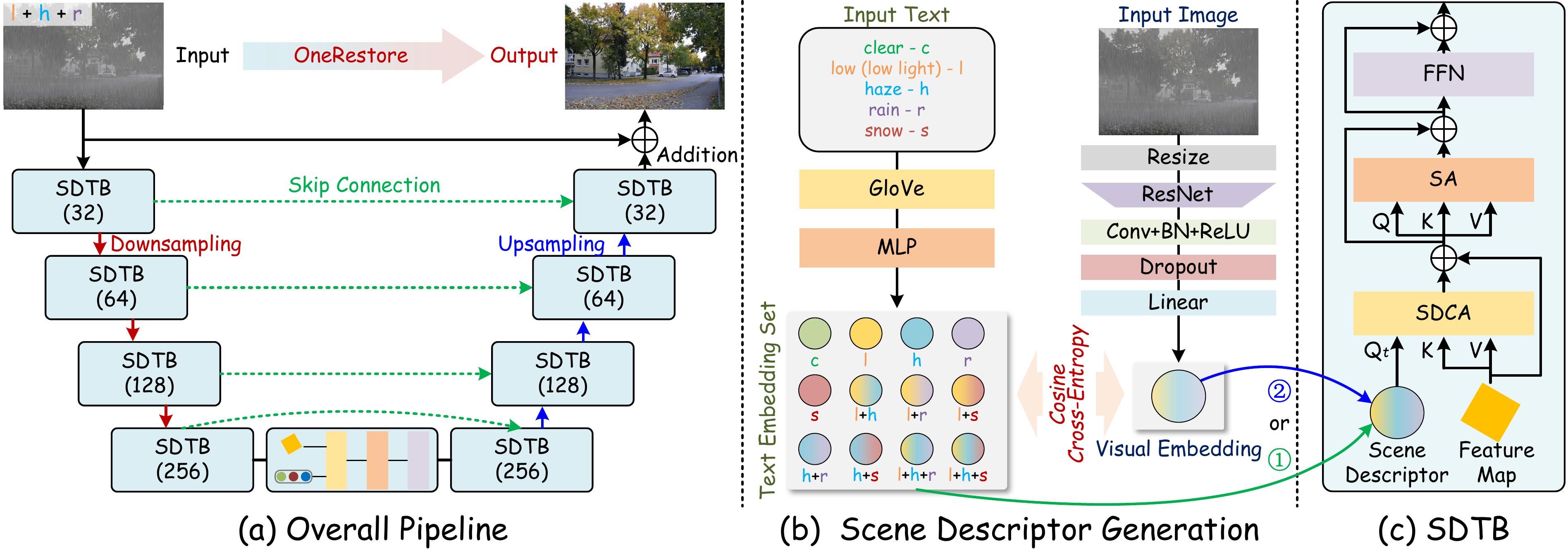}\\
        \vspace{-2mm}
        \caption{Architecture of our OneRestore. (a) Overall pipeline, where 32, 64, 128, and 256 represent the number of channels. (b) Scene descriptor generation, where scene descriptors are fed into each (c) Scene Descriptor-guided Transformer Block (SDTB) by manual text embeddings or automatic extraction based on visual attributes.}
        \vspace{-2mm}
        \label{fig:OneRestore}
    \end{figure*}
\section{OneRestore}
\label{sec:network}
\subsection{Overall Pipeline}
    The encoder-decoder network, validated as an effective architectural paradigm for image restoration~\cite{guo2023scanet, chen2023learning, du2020blind}, serves as the overall structure for our model, as depicted in Fig.~\ref{fig:OneRestore}a. Building upon the encoder-decoder structure, we devise an innovative Scene Descriptor-guided Transformer Block (SDTB) as the basic unit for feature extraction (Fig.~\ref{fig:OneRestore}c). Our approach diverges from existing transformer-based methods~\cite{chen2023msp, chen2023learning, song2023vision} by concurrently integrating image features and scene descriptors into the SDTB basic block for supporting fine-grained image restoration. In particular, the generation of scene descriptors ranges from manual text embeddings to automatic extraction based on visual attributes, as illustrated in Fig.~\ref{fig:OneRestore}b. Further details on scene descriptor generation and SDTB are elaborated in Sections~\ref{sec:weg} and~\ref{sec:web}. For capturing multi-scale representations, the encoder employs three downsampling operations, mirrored by the decoder's upsampling operations. To be specific, we utilize "Max-Pooling + Conv" and "Bilinear Interpolation + Conv" for downsampling and upsampling, respectively. Simultaneously, three skip connections and a global residual enhance model convergence by bridging shallow and deep features. Additional details regarding the configuration of our OneRestore can be found in \textbf{supplementary materials}.
    Ultimately, the overall loss $\mathcal{L}$ for model training is defined as follows
    \begin{equation}
        \label{eq:loss}
        \mathcal{L} = \alpha_1 \mathcal{L}^s_{1}(J,\hat{J})+ \alpha_2 \mathcal{L}_\text{M}(J,\hat{J})+ \alpha_3\mathcal{L}_\text{c}(J,\hat{J},I,\{I_o\}),
    \end{equation}
    where $\mathcal{L}^s_1$, $\mathcal{L}_\text{M}$, and $\mathcal{L}_\text{c}$ represent the smooth $l_1$ loss~\cite{girshick2015fast}, Multi-Scale Structural Similarity (MS-SSIM) loss, and the proposed composite degradation restoration loss (see Sec.~\ref{sec:sdrl}), respectively, $\alpha_{1\text{-}3}$ denote the penalty coefficients, $J$, $\hat{J}$, $I$, and $\{I_o\}$ are the positive, model output anchor, model input negative, and other negatives, respectively.
\subsection{Scene Descriptor Generation}
\label{sec:weg}
    To augment the controllability of our OneRestore, we introduce the scene description embedding as an additional input to the model, alongside the degraded image. As delineated in Fig.~\ref{fig:OneRestore}b, the generation of the scene descriptor offers two selectable modes: the manual mode {\color{green}\ding{172}} and the automatic mode {\color{blue}\ding{173}}. In the manual mode, user involvement is necessitated to supply scene description text, generating the corresponding text embedding employed as an input. Conversely, the automatic mode involves the extraction of visual attributes from the image to generate visual embedding, estimating the most proximate text embedding. To quantify the difference between visual and text embeddings and train optimal text and visual embedders, we introduce cosine cross-entropy loss for model weight optimization.

    \textbf{Text Embedder.}
    For the text embedding generation task, we employ a set of 5 scene description texts as input to generate 12 text embeddings. Initially, GloVe~\cite{pennington2014glove} is leveraged to generate initial text embeddings for 1 clear scene and 4 scenes with single degradation types. Subsequently, 7 additional composite degradation text embeddings (i.e., \textit{low+haze}, \textit{low+rain}, \textit{low+snow}, \textit{haze+rain}, \textit{haze+snow}, \textit{low+haze+rain}, and \textit{low+haze+snow}) are generated by averaging the element values in the corresponding text embeddings. A Multi-Layer Perception (MLP) is then employed to refine the 12 text embeddings.

    \textbf{Visual Embedder.}
    Taking inspiration from the work of Saini et al.~\cite{saini2022disentangling}, we first resize the input image to 224$\times$224 and extract image features using a ResNet-18~\cite{he2016deep} network without average pooling. The initial weights of ResNet-18 are pre-trained on the ImageNet dataset. Subsequently, a single convolutional layer, followed by a dropout layer and a linear layer, is introduced to generate the final visual embedding. By calculating the cosine similarity between visual embeddings and all text embeddings, and selecting the one with the highest similarity, we can automatically generate scene descriptors to achieve image restoration based on visual attribute control.

    \textbf{Cosine Cross-Entropy.}
    Similar to~\cite{saini2022disentangling}, the score of visual embedding and each text embedding is obtained by ``Cosine Similarity + Softmax''. Given $e_v$ and $e_t$ being the visual and text embeddings, the similarity score $S(e_v, e_t)$ can be given by
    \begin{equation}
    \begin{split}
    &cos(e_v, e_t) = \delta \cdot \frac{e_v \cdot e^\top _t}{\left \|  e_v\right \|\left \|  e_t\right \| }, \label{cos} \\
    &S(e_v, e_t) = \frac{e^{cos(e_v, e_t)}}{ {\textstyle \sum_{t_i=1}^{N_t}}e^{cos(e_v,e_{t_i})}}, \\
    \end{split}
    \end{equation}
    with $\delta$ and $N_t$ being the temperature factor and the number of text embeddings, respectively. Based on the calculated $S(e_v, e_t)$, we use the cross-entropy loss for model training.
\subsection{Scene Descriptor-guided Transformer Block}
\label{sec:web}
    To further harness the capabilities of scene description embeddings in image restoration tasks and facilitate a more adaptive and nuanced restoration process, we introduce the innovative Scene Descriptor-guided Transformer Block (SDTB), illustrated in Fig.~\ref{fig:OneRestore}c. The SDTB consists of three primary components, namely Scene Descriptor-guided Cross-Attention (SDCA), Self-Attention (SA), and Feed-Forward Network (FFN). Diverging from traditional self-attention mechanisms, the proposed SDCA module utilizes a query generated from scene descriptors instead of queries generated from image features. The mathematical representation of our SDCA can be written as
    \begin{equation}
        \label{eq:ca}
        \text{SDCA}(\mathbf{Q}_t, \mathbf{K}, \mathbf{V}) = \text{Softmax}(\frac{\mathbf{Q}_t \cdot \mathbf{K}^\top}{\lambda})\mathbf{V},
    \end{equation}
    where $\lambda$ is a temperature factor, $\mathbf{Q}_t$ represents the scene descriptor-generated query matrix, $\mathbf{K}$ and $\mathbf{V}$ denote the image feature-generated key and value matrices, respectively. $\mathbf{Q}_t$ is obtained from the input scene description embedding $e_t$ via a linear layer. During the process of matrix multiplication between $\mathbf{Q}_t$ and $\mathbf{K}$. To ensure consistency in the number of tokens between $\mathbf{K}$ and $\mathbf{Q}_t$, we resize the original image before performing the reshaping operation to generate $\mathbf{K}$. By incorporating the SDCA module, each transformer block in OneRestore can effectively integrate scene descriptors and image features for targeted image restoration. As for the SA and FFN modules, we adopt structures akin to those in Restormer~\cite{zamir2022restormer}. More details of our SDTB are introduced in the \textbf{supplement}.

\subsection{Composite Degradation Restoration Loss}
\label{sec:sdrl}
\begin{figure}[t]
        \centering
        \includegraphics[width=1\linewidth]{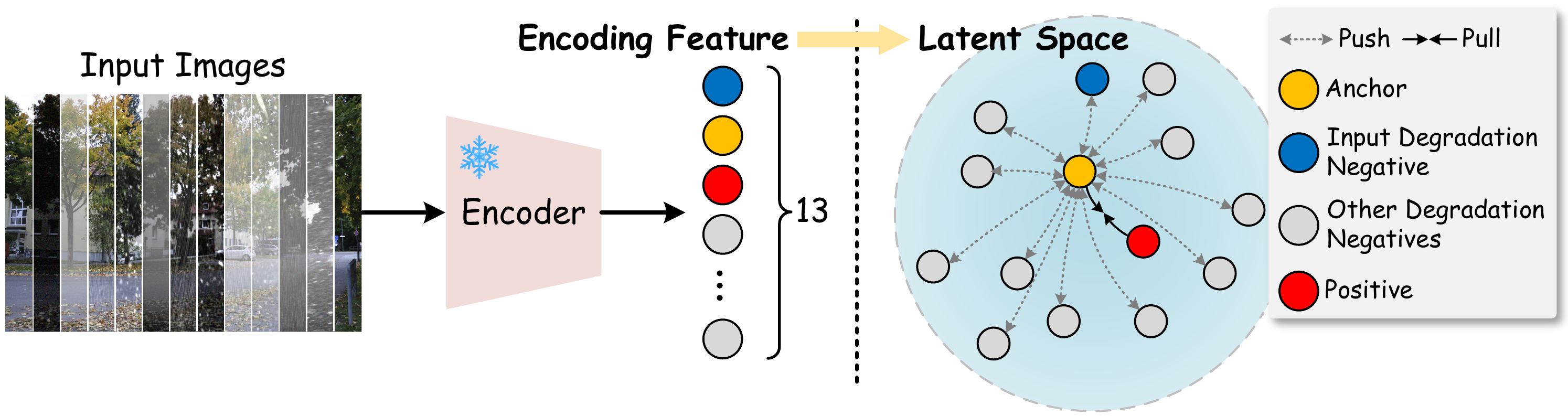}
        \vspace{-6mm}
        \caption{Illustration of the composite degradation restoration loss.}
        \vspace{-4mm}
        \label{fig:conloss}
    \end{figure}

    Using contrastive loss has been demonstrated as an effective way of improving image restoration performance~\cite{wu2021contrastive, zheng2023curricular}. Essentially, contrastive loss views the degraded input image as a negative sample, the clear image as a positive sample, and the restored image as an anchor. This loss function works to minimize the distance between the anchor and the positive sample while maximizing its distance from the negative sample. In the composite degradation image restoration task, the traditional contrastive loss, however, may inadvertently reduce the distance between the anchor and other degraded images. To make the anchor closer to the positive sample while maintaining a clear separation from all degraded negatives, we design a composite degradation restoration loss, illustrated in Fig.~\ref{fig:conloss}. Considering $I$, ${I_o}$, $J$, and $\hat{J}$ as the input negative, other negatives, positive, and anchor, the proposed loss can be written as
    \begin{footnotesize}
    \begin{equation}
    \label{eq:contr}
    \mathcal{L}_\text{c}(J,\hat{J},I,\{I_o\}) = \sum_{k=1}^{K} \xi_k\frac{\mathcal{L}_1(V_k(J),V_k(\hat{J}))}{\xi_c\mathcal{L}_1(V_k(\hat{J}),V_k(I)) + \sum_{o=1}^{O}\xi_o\mathcal{L}_1(V_k(I_o),V_k(\hat{J}))},
    \end{equation}
    \end{footnotesize}
    where $V_k(\cdot)$ denotes the extraction operation of $k$-th hidden feature from the VGG-16 pre-trained on the ImageNet, $\mathcal{L}_1$ represents the $l_1$ loss, $K$ and $O$ are the numbers of used VGG-16 feature layers and other degraded negatives, $\xi_k$, $\xi_c$, and $\xi_o$ denote the hyper-parameters, respectively. In this work, the number of other degraded negatives $O$ is $10$. We use the output features of 3-rd, 8-th, and 15-th layers from VGG-16 and set $K=3$. Furthermore, we consider that the features of each feature layer of each degraded negative have equal importance. Therefore, we set $\xi_c=\xi_o=\frac{1}{11}$ and $\xi_k=\frac{1}{3}$. According to our ablation study, the proposed loss confers a significant advantage over using a single input degradation as a negative sample, which can easily capture positive sample features.

    \setlength{\tabcolsep}{25pt}
    \begin{table}[h]
    \centering
    \vspace{-4mm}\caption{Settings of model training. ``lr'' denotes the learning rate.}
    \vspace{-2mm}
    \label{tb:train}
    \begin{tabular}{l|c|c}
        \hline \hline
        Terms           & Text/Visual Embedder & OneRestore       \\ \hline
        Epoch           & 200                  & 120              \\ \hline
        Initial lr      & 0.0001               & 0.0002           \\ \hline
        lr Decay        & 0.5/50 epochs        & 0.5/20 epochs    \\ \hline
        Batch Size      & 256                  & 4                \\ \hline \hline
    \end{tabular}
    \vspace{-12mm}
    \end{table}
    
\section{Experiments}
\label{sec:exp}
\subsection{Experiment Settings}
\textbf{Implementation Details.}
    Our OneRestore is implemented by PyTorch 1.12.0 and trained on a PC with 2 AMD EPYC 7543 32-Core Processors and 8 NVIDIA L40 GPUs. We use the Adam with exponential decay rates being $\beta_1=0.9$ and $\beta_2=0.999$ for optimization. The training parameter settings for text/visual embedder and OneRestore are shown in Table~\ref{tb:train}. In particular, all images used for training are cropped into 256$\times$256 image patches with a sampling stride of 200 and randomly flipped by 0, 90, 180, and 270 degrees to generate 312k patch pairs. More details about training and inference can be found in the \textbf{supplement}.
    
\textbf{Evaluation Datasets.}
    To facilitate a thorough comparison, we execute image restoration experiments across various datasets. These include the Composite Degradation Dataset (CDD-11) crafted by us, established One-to-One benchmarks, and authentic real-world datasets. The standard One-to-One benchmarks are designated for four principal tasks: the LOw-Light dataset (LOL)~\cite{Chen2018Retinex}, the REalistic Single Image DEhazing Outdoor Training Set (RESIDE-OTS)~\cite{li2018benchmarking}, the Rain1200 dataset~\cite{zhang2018density}, and the Snow100k dataset~\cite{liu2018desnownet}. Owing to the dearth of specialized datasets for real-world composite degradation, we have curated a selection from the LIME low-light enhancement dataset~\cite{guo2016lime}, the RESIDE Real-world Task-driven Testing Set (RESIDE-RTTS)~\cite{li2018benchmarking}, Yang's rainy dataset \cite{yang2017deep}, and the realistically snowy Snow100k dataset~\cite{li2018benchmarking}. 
    %
    \setlength{\tabcolsep}{3pt}
    \begin{table}[t]
        \centering
        \caption{Comparison of quantitative results on CDD-11 dataset. OneRestore$^\dagger$ means using the corresponding scene description text as additional input, and WGWSNet requires the scene type as model input. {\color{red} Red}, {\color{green} green}, and {\color{blue} blue} indicate the best, second-best, and third-best results, respectively.}
        \label{tb:syn}
        \begin{tabular}{c|l|c|c|c|c}
            \hline \hline
            Types & Methods                                        & Venue \& Year & PSNR $\uparrow$& SSIM $\uparrow$                                 & \#Params \\ \hline
            & Input                                          &             & 16.00 & 0.6008                               & -        \\ \hline
            \multirow{9}{*}{One-to-One} & MIRNet  ~\cite{zamir2020learning}              & ECCV2020    & 25.97 & 0.8474                               & 31.79M   \\
            \multirow{9}{*}{} & MPRNet~\cite{zamir2021multi}                   & CVPR2021    & 25.47 & 0.8555                               & 15.74M   \\
            \multirow{9}{*}{} & MIRNetv2~\cite{zamir2022learning}              & TPAMI2022   & 25.37 & 0.8335                               & 5.86M    \\
            \multirow{9}{*}{} & Restormer~\cite{zamir2022restormer}            & CVPR2022    & 26.99 & {\color{blue}0.8646}                 & 26.13M   \\
            \multirow{9}{*}{} & DGUNet~\cite{mou2022deep}                      & CVPR2022    & 26.92 & 0.8559                               & 17.33M   \\
            \multirow{9}{*}{} & NAFNet~\cite{chen2022simple}                   & ECCV2022    & 24.13 & 0.7964                               & 17.11M   \\
            \multirow{9}{*}{} & SRUDC~\cite{song2023under}                     & ICCV2023    & {\color{blue}27.64} & 0.8600                 & 6.80M    \\
            \multirow{9}{*}{} & Fourmer~\cite{zhou2023fourmer}                     & ICML2023    & 23.44 & 0.7885                 &  0.55M   \\ 
            \multirow{9}{*}{} & OKNet~\cite{cui2024omni}                     & AAAI2024    & 26.33 & 0.8605                 &   4.72M  \\ \hline
            \multirow{4}{*}{One-to-Many} & AirNet~\cite{li2022all}                        & CVPR2022    & 23.75 & 0.8140                               & 8.93M    \\
            \multirow{4}{*}{} & TransWeather~\cite{valanarasu2022transweather} & CVPR2022    & 23.13 & 0.7810                               & 21.90M   \\
            \multirow{4}{*}{} & WeatherDiff~\cite{ozdenizci2023restoring}      & TPAMI2023   & 22.49 & 0.7985                                        & 82.96M   \\
            \multirow{5}{*}{} & PromptIR~\cite{potlapalli2024promptir}      & NIPS2023   & 25.90 & 0.8499                                        & 38.45M   \\
            \multirow{4}{*}{} & WGWSNet~\cite{zhu2023learning}                 & CVPR2023    & 26.96 & 0.8626                               & 25.76M   \\ \hline
            \multirow{2}{*}{One-to-Composite} & OneRestore                                     &             & {\color{green}28.47}&{\color{green}0.8784} & 5.98M    \\ 
            & OneRestore$^\dagger$                           &             & {\color{red}28.72}&{\color{red}0.8821}     & 5.98M    \\ \hline \hline
        \end{tabular}
        \vspace{-4mm}
    \end{table}

\textbf{Competitors and Evaluation Metrics.}
    We compare our OneRestore with 9 One-to-One image restoration methods (MIRNet~\cite{zamir2020learning}, MPRNet~\cite{zamir2021multi}, MIRNetv2~\cite{zamir2022learning}, Restormer~\cite{zamir2022restormer}, DGUNet~\cite{mou2022deep}, NAFNet~\cite{chen2022simple}, SRUDC~\cite{song2023under}, Fourmer~\cite{zhou2023fourmer}, and OKNet~\cite{cui2024omni}) and 5 One-to-Many image restoration methods (AirNet~\cite{li2022all}, TransWeather~\cite{valanarasu2022transweather}, WeatherDiff~\cite{ozdenizci2023restoring}, PromptIR~\cite{potlapalli2024promptir}, and WGWSNet~\cite{zhu2023learning}). In particular, all compared state-of-the-art methods are retrained on our CDD-11 train set. For PromptIR and WGWSNet, we adjust the number of prompts and the weather-specific parameters of each layer for corresponding to our task, respectively. Furthermore, we adopt the Peak Signal-to-Noise Ratio (PSNR) and Structure Similarity Index Measure (SSIM) as evaluation metrics.
    
\subsection{Comparison with SOTAs}

    \begin{wrapfigure}{r}{6cm}
        \centering
        \vspace{-8mm}
        \includegraphics[width=0.5\textwidth]{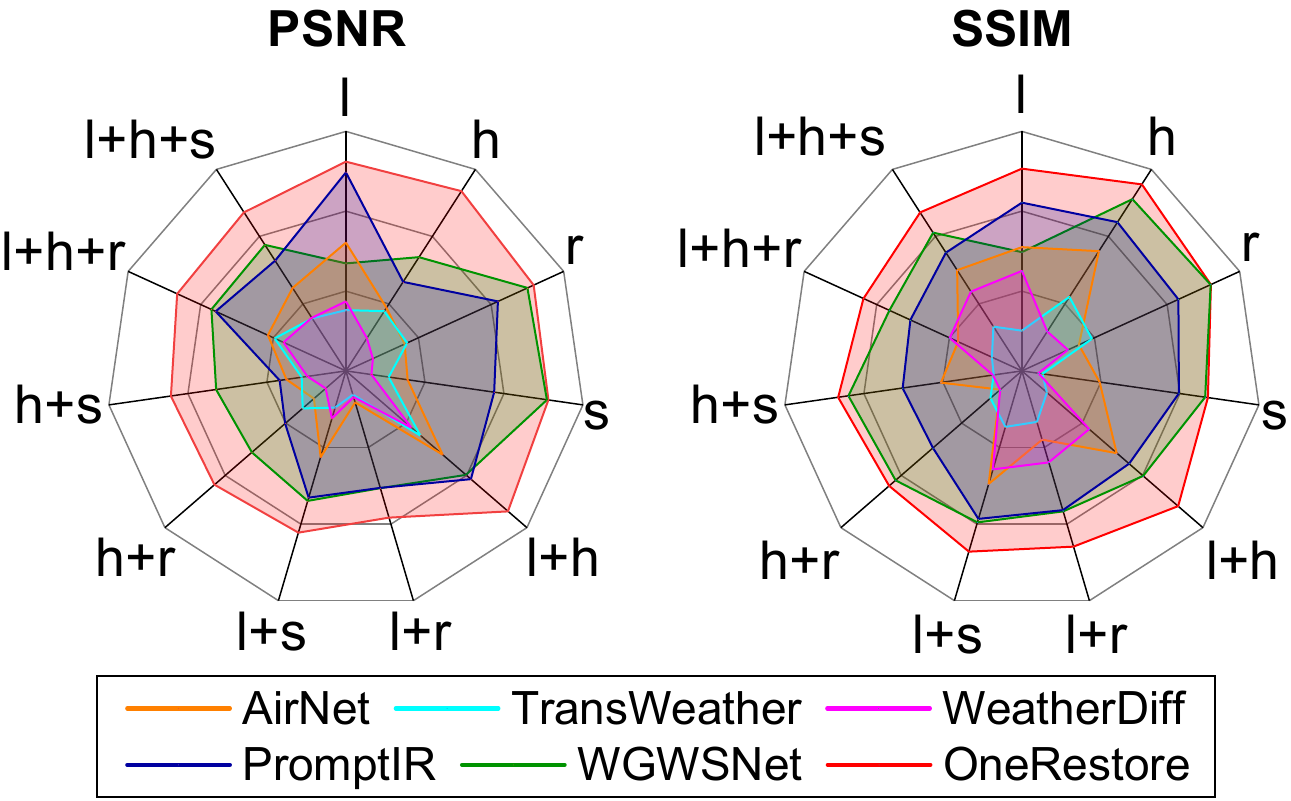}
        \vspace{-4mm}
        \caption{Comparison of quantitative results for different degradation scenarios on CDD-11 dataset.}
        \vspace{-6mm}
        \label{fig:metric}
    \end{wrapfigure}
\begin{figure*}[t]
        \centering
        \includegraphics[width=1\linewidth]{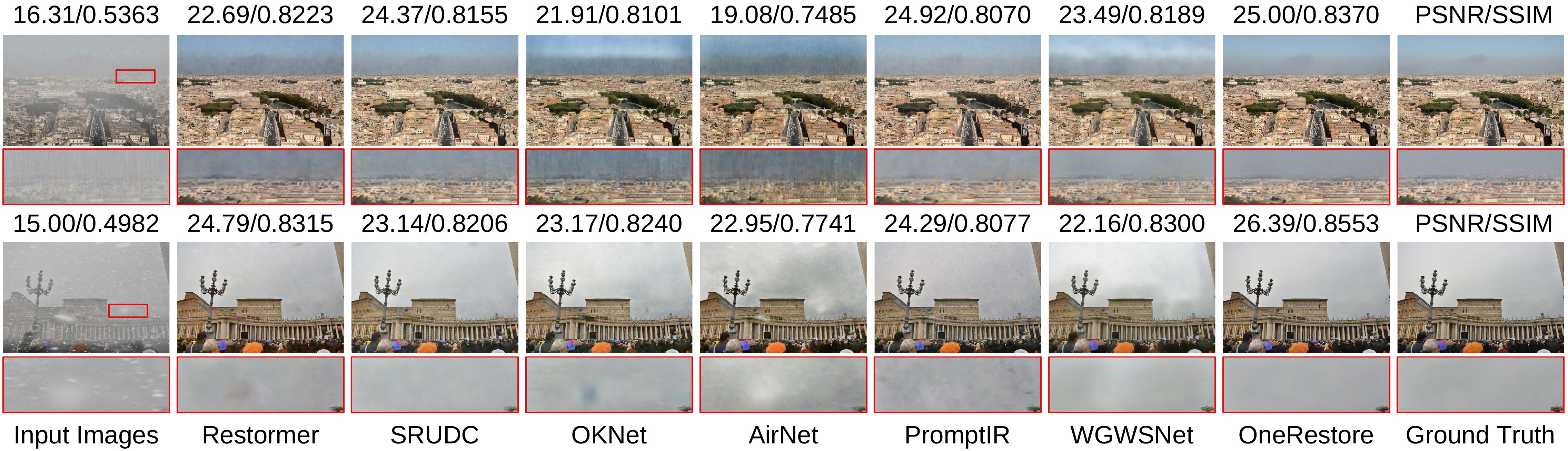}
        \vspace{-6mm}
        \caption{Comparison of image restoration on \textit{low+haze+rain} (top) and \textit{low+haze+snow} (bottom) synthetic samples.}
        \label{fig:enhance_syn}
        \vspace{-2mm}
    \end{figure*}

    The quantitative evaluation results on our CDD-11 test dataset are displayed in Table~\ref{tb:syn}. OneRestore$^\dagger$ with text embedder achieves the highest performance, followed by OneRestore with visual embedder as the second-best. Both OneRestore$^\dagger$ and OneRestore demonstrate substantial leads over the third-ranked methods in terms of both PSNR and SSIM. Further detailed restoration performance comparison of our method and the One-to-Many methods in each degradation scenario is illustrated in Fig.~\ref{fig:metric} using radar charts. Notably, image restoration for composite degradation scenarios proves challenging. By incorporating scene description embeddings into image restoration models and utilizing the proposed composite degradation restoration loss, our method robustly reconstructs clear images across various situations. For compelling evidence, Fig.~\ref{fig:enhance_syn} provides the visual comparison of image restoration in two composite degradation samples. It is evident that SRUDC, OKNet, AirNet, and WGWSNet are unable to suppress all degenerative interferences, while Restormer and PromptIR are prone to color distortion and noise residual. In contrast, the proposed OneRestore can produce more natural results and fully preserve image textures and details.

\begin{figure}[t]
        \centering
        \includegraphics[width=1\linewidth]{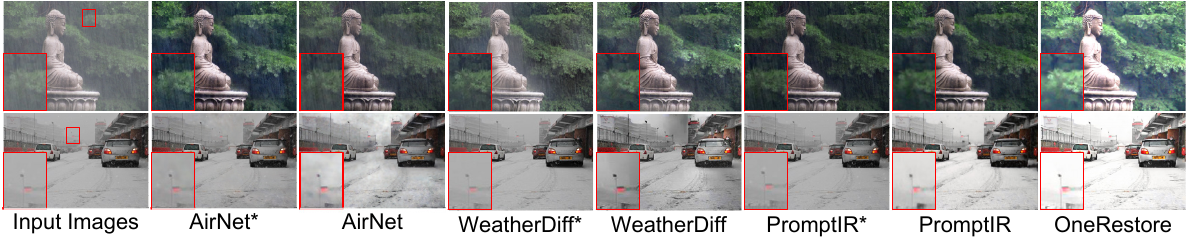}
        \vspace{-6mm}
        \caption{Comparison of image restoration on \textit{low+haze+rain} (top) and \textit{low+haze+snow} (bottom) samples in real-world scenarios. * represents the utilization of original weights published in the author's code.}
        \vspace{-2mm}
        \label{fig:enhance_real}
    \end{figure}

    Moreover, Fig.~\ref{fig:enhance_real} displays the image restoration results in two real-world scenarios. Compared with existing One-to-Many image restoration methods~\cite{li2022all, ozdenizci2023restoring, potlapalli2024promptir}, the proposed OneRestore can simultaneously focus on multiple degradation factors and robustly reconstruct scene details. More results on CDD-11, real-world scenarios, and other One-to-One benchmarks of low-light, hazy, rainy, and snowy scenes are included in \textbf{supplementary materials}.
\subsection{Ablation Study}
\label{sec:ab}

    \textbf{Effectiveness of Network Modules.}
    Table~\ref{tb:ab_module1} reports the quantitative evaluation of different module configurations. While the Feed-Forward Network (FFN) with convolution as the basic unit serves as a crucial module in the transformer network, it alone fails to generate satisfactory reconstruction results. The ordinary Self-Attention module (SA) significantly strengthens network performance, but this method regrets that the model fails to be flexibly controlled. Compared to depending only on SA, relying exclusively on Scene Description-guided Cross-Attention (SDCA) utilizing scene description embedding yields better outcomes. This further confirms that addressing complex composite degradations can be done more effectively and flexibly by integrating degradation scene descriptors into the model. Ultimately, the best performance can be achieved by using a combination of SDCA, SA, and FFN.

    \begin{wrapfigure}{r}{6cm}
        \centering
        \vspace{-8mm}
        \includegraphics[width=0.5\textwidth]{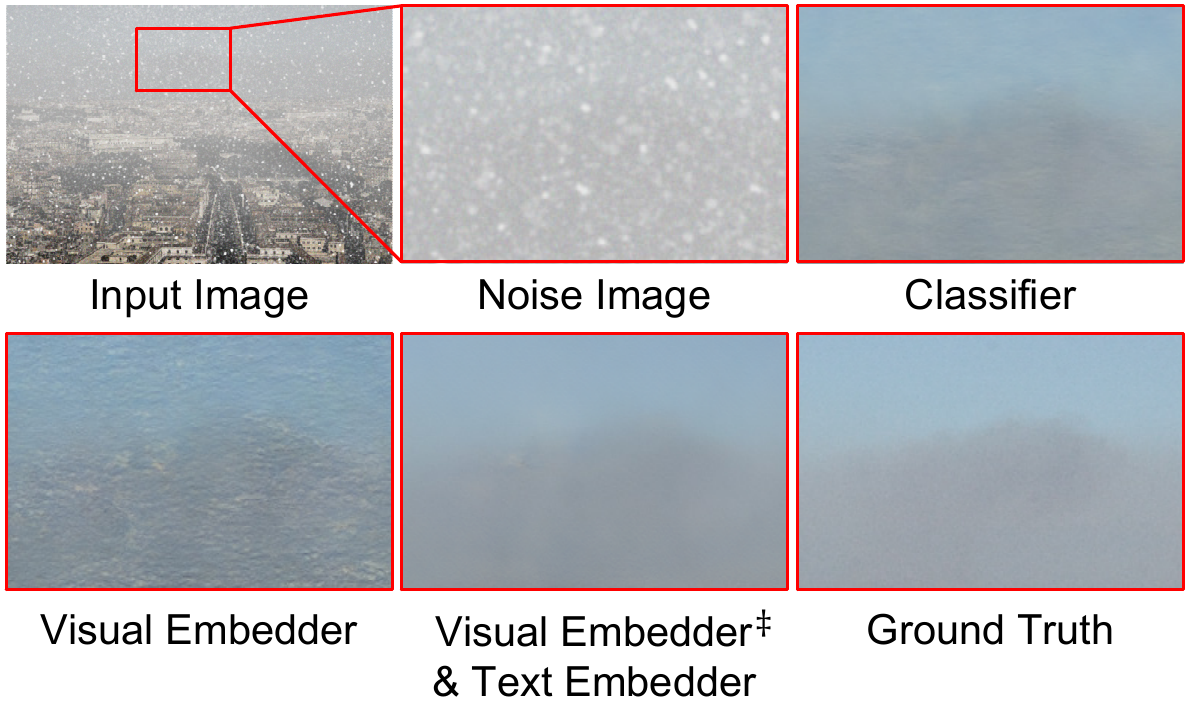}
        \vspace{-7mm}
        \caption{Comparison of image restoration on different embedding strategies.}
        \vspace{-6mm}
        \label{fig:AB}
    \end{wrapfigure}
    \textbf{Selection of Description Embedding Strategies.}
    In this section, we compare the image restoration performance by using different strategies to generate scene descriptor, including Classifier (utilizing the embedding before the last linear layer of the retrained AlexNet~\cite{krizhevsky2017imagenet} as the scene descriptor), Visual Embedder (employing visual embedding for model training), Visual Embedder$^\ddagger$ (deriving text embedding via visual embedder), and Text Embedder  (employing text embedding for model training). The quantitative evaluation results are presented in Table~\ref{tb:ab_module2}. It is observed that employing Classifier and Visual Embedder to generate scene descriptors yields comparable outcomes. However, these approaches have a drawback in that each image has a distinct embedding as input, making it challenging for the model to recognize the cause of degradation. Fig. \ref{fig:AB} displays a classic case comparison using different embedding strategies, where the abnormal artifacts of the results produced by Classifier and Visual Embedder are caused by the model's fuzzy identification of degradation factors. Moreover, the restoration model lacks controllability. The discrepancy between Visual Embedder$^\ddagger$ and Text Embedder stems from errors in scene descriptor estimation. Visual Embedder$^\ddagger$ has an accuracy of 97.55\% on the test dataset, with misestimation occurring when some degradation types are insignificant. For instance, a \textit{haze+snow} scene may be evaluated as a \textit{haze} or \textit{snow} scene. In contrast, the text embedding-based approach is superior in describing degradation scenes and recovering images, using a fixed number of scene descriptors. This allows the model to employ degraded scene descriptors as switches to achieve controlled restoration. 
    \setlength{\tabcolsep}{14pt}
    \begin{table*}[t]
        \centering
        \caption{Ablation study for different model configurations. SDCA, SA, and FFN denote the scene description-guided cross-attention, self-attention, and feed-forward network, respectively.}
        \vspace{-2mm}
        \label{tb:ab_module1}
        \begin{tabular}{ccc|c|c|c}
            \hline \hline
            SDCA         & SA         & FFN        & PSNR $\uparrow$ & SSIM $\uparrow$            & Controllability \\ \hline
                       &            & \checkmark & 24.81 & 0.8607          &                 \\
                       & \checkmark & \checkmark & 27.19 & 0.8697        &                 \\
            \checkmark &            & \checkmark & 27.93 & 0.8767         & \checkmark      \\
            \rowcolor{lightgray} \checkmark & \checkmark & \checkmark & \textbf{28.72} & \textbf{0.8821} & \checkmark      \\ \hline \hline
        \end{tabular}
        \vspace{-2mm}
    \end{table*}
    \begin{figure*}[t]
        \centering
        \includegraphics[width=1\linewidth]{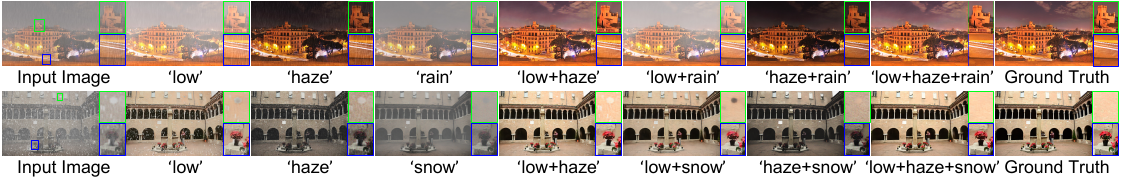}
        \vspace{-5mm}
        \caption{Comparison of image restoration on \textit{low+haze+rain} (top) and \textit{low+haze+snow} (bottom) synthetic samples by using different texts.}
        \label{fig:control}
        \vspace{-4mm}
    \end{figure*}

    \setlength{\tabcolsep}{19pt}
    \begin{table}[t]
        \centering
        \caption{Ablation study for description embedding strategies.}
        \vspace{-2mm}
        \label{tb:ab_module2}
        \begin{tabular}{c|c|c|c}
            \hline \hline
            Models                     & PSNR $\uparrow$&SSIM $\uparrow$             & Controllability \\ \hline
            Classifier         & 28.19&0.8783          &                 \\
            Visual Embedder            & 28.24& 0.8781          &                 \\
            \rowcolor{lightgray} Visual Embedder$^\ddagger$ & 28.47 & 0.8784         &                 \\
            \rowcolor{lightgray} Text Embedder              & \textbf{28.72} & \textbf{0.8821} & \checkmark      \\ \hline \hline
        \end{tabular}
        \vspace{-2mm}
    \end{table}
    \setlength{\tabcolsep}{14pt}
    \begin{table}[t]
        \centering
        \caption{Ablation study for different loss functions. CL and CDRL denote the ordinary contrastive loss and the proposed composite degradation restoration loss, respectively.}
        \vspace{-2mm}
        \label{tb:ab_module3}
        \begin{tabular}{cccc|c|c}
            \hline \hline
            Smooth $l_1$  & MS-SSIM    & CL         & CDRL          & PSNR $\uparrow$ & SSIM $\uparrow$            \\ \hline
            \checkmark &            &            &               &  28.16 &    0.8633        \\
            \checkmark & \checkmark &            &               & 27.54 & 0.8708          \\
            \checkmark & \checkmark & \checkmark &               &  27.61 &  0.8723          \\
            \rowcolor{lightgray} \checkmark & \checkmark &            & \checkmark    & \textbf{28.72} & \textbf{0.8821} \\ \hline \hline
        \end{tabular}
    \vspace{-4mm}
    \end{table}
    \begin{wrapfigure}{r}{6cm}
        \centering
        \vspace{-8mm}
        \includegraphics[width=0.5\textwidth]{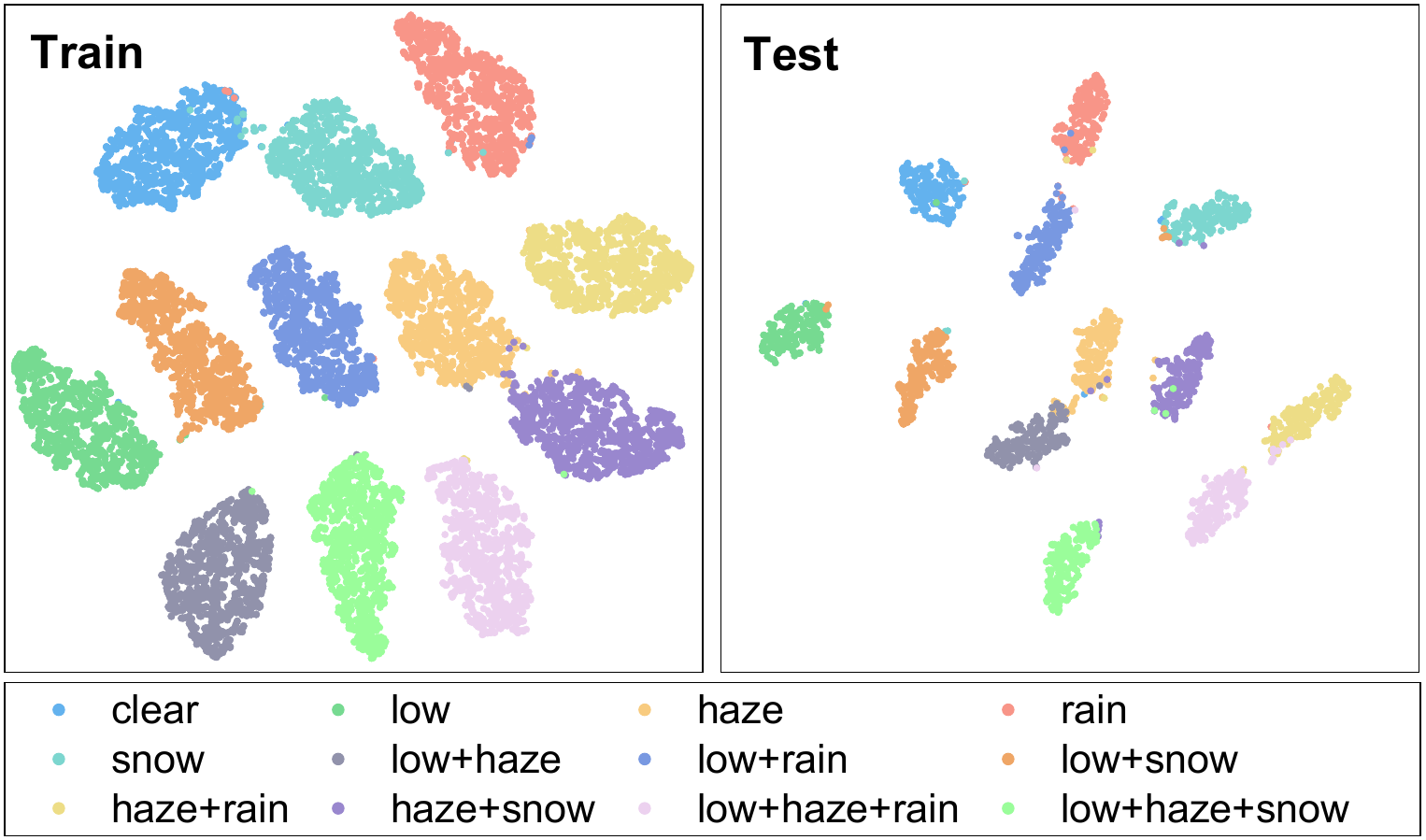}
        \vspace{-6mm}
        \caption{Visualization of visual embeddings based on t-SNE.}
        \vspace{-6mm}
        \label{fig:tsne}
    \end{wrapfigure}
    \textbf{Effectiveness of Loss Functions.} 
    Table~\ref{tb:ab_module3} presents a performance comparison of different loss functions. Notably, the integration of the MS-SSIM loss is instrumental in elevating SSIM scores at the cost of reduced PSNR performance. In contrast, the CL loss demonstrates minimal impact, providing only marginal improvements in SSIM and PSNR. Upon replacing CL with CDRL, a substantial enhancement is observed in both PSNR and SSIM. The proposed CDRL loss yields an output that closely resembles the clear image while avoiding other forms of degradation.

\subsection{Model Control via Scene Descriptor}
\label{sec:control}

    To demonstrate the controllability of the scene descriptor in our OneRestore, we conduct experiments on two intricate synthetic degradation scenarios, as depicted in Fig.~\ref{fig:control}. By manually adjusting specific text to generate corresponding degraded scene description embeddings, our model can selectively focus on different degradation factors, achieving targeted restorations. Our approach excels in controllable restoration without the need for multiple sets of weight screening, making it more concise compared to other One-to-Many methods that rely on partial parameter sharing. Meanwhile, our method can support adaptive restoration through the estimation of text embeddings based on visual attribute extraction. Fig.~\ref{fig:tsne} employs the t-SNE~\cite{van2008visualizing} to visualize the embeddings generated by the visual embedder on the CDD-11 train and test sets. Obviously, the constructed visual embedder can clearly distinguish between different scenarios. Additionally, more real cases of the model control using different scene description texts are presented in \textbf{supplementary materials}.
\section{Conclusion}
\label{sec:con}

    Our proposed OneRestore framework represents a significant step forward in image restoration, synthesizing a range of degradation patterns to accurately simulate complex scenarios. The innovative use of a scene descriptor-guided cross-attention block within a transformer-based model facilitates an adaptable and fine-grained restoration process, yielding results that surpass prior methods in both synthetic and real-world datasets. However, despite its advancements, OneRestore exhibits limitations in processing extremely high-density corruption scenarios and high-complex corruption scenarios containing unconsidered degradations, where the model's predictive capabilities can be challenged. Future work will aim to enhance the robustness of OneRestore against such extreme conditions and to reduce computational overhead, further extending its applicability and efficiency. Our findings underscore the potential of universal scene restoration, with the caveat that continued refinement is essential for tackling the full spectrum of real-world image degradation.

\textbf{Acknowledgement.}
    This project is supported by the Guangdong Natural Science Funds for Distinguished Young Scholar under Grant 2023B1515020097 and National Research Foundation Singapore under the AI Singapore Programme under Grant AISG3-GV-2023-011.
%
%
\bibliographystyle{splncs04}
\bibliography{main}
\newpage

\appendix
\section{Overview}
    This supplementary material provides more details on model configuration and experimental results, which can be listed as follows:
    \begin{itemize}
        \item We offer additional configuration details about the Text/Visual Embedder and the OneRestore model.
        \item We elaborate on model training and inference procedures.
        \item We present more experimental results to verify the effectiveness and controllability of the proposed method.
        \item We discuss limitations and outline future research directions.
    \end{itemize}
    \begin{figure}[h]
        \centering
        \includegraphics[width=1\linewidth]{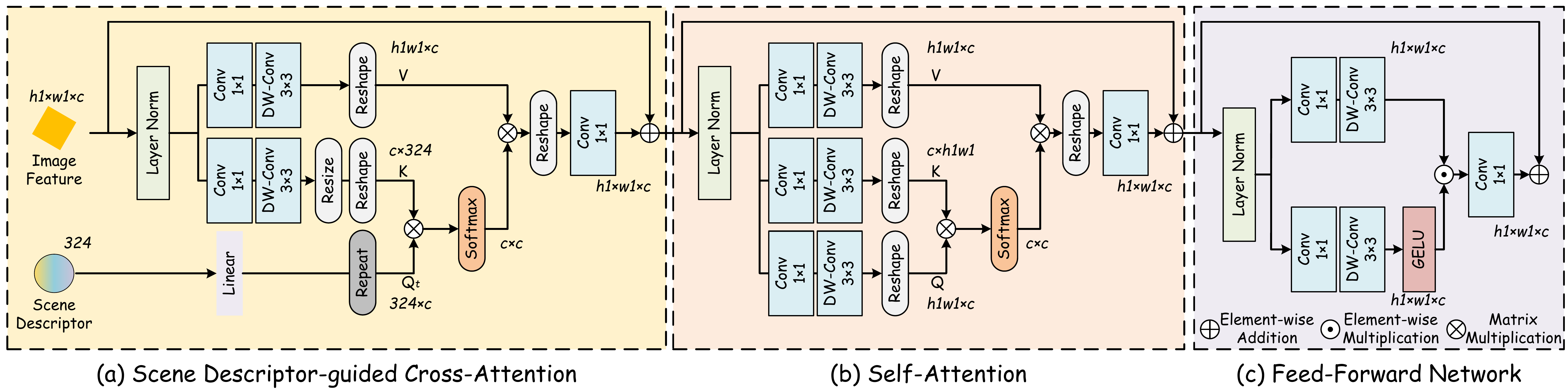}
        \vspace{-6mm}
        \caption{Architecture of proposed Scene Descriptor-guided Transformer Block (SDTB).}
        \vspace{-10mm}
        \label{fig:transformer}
    \end{figure}
\section{Network Details}
\subsection{Transformer Block}
    Fig.~\ref{fig:transformer} displays the details of the Scene Descriptor-guided Transformer Block (SDTB), which includes three parts: Scene Descriptor-guided Cross-Attention (SDCA), Self-Attention (SA), and Feed-Forward Network (FFN). Specifically, the SDCA considers the image feature and degraded scene descriptor as input. The values $\mathbf{V}$ and keys $\mathbf{K}$ are generated by processing image features using two modules with the same structure but different weights, concatenating $1 \times 1$ Convolution (Conv) with $3 \times 3$ DepthWise Convolution (DW-Conv). As for the scene descriptor, we use a linear layer to produce the scene description query $\mathbf{Q}_t$. Then, we adjust the image size to make the number of tokens in $\mathbf{K}$ to be consistent with $\mathbf{Q}_t$. The SA and FFN adopt modules proposed by Restormer\cite{zamir2022restormer}. By incorporating SDCA, our model can facilitate restored orientation control by changing different embeddings.
    \setlength{\tabcolsep}{6pt}
    \begin{table}[t]\scriptsize
        \centering
        \caption{Configurations of network architecture. $n$, $c$, and $head$ denote the numbers of texts, channels, and attention heads, respectively. $l$ is the length of the initial text embedding. $k$ and $s$ represent the kernel size and stride, respectively. $bn$ and $relu$ denote the batch normalization and ReLU activation function, respectively.}
        \vspace{-3mm}
        \label{tb:config}
        \begin{tabular}{c|l|l|l}
            \hline \hline 
            Models                                                                    & Layers   & Configurations & Output Size                        \\  \hline
            \multirow{3}{*}{Text Embedder}   & Input1   & Scene Description Text & $1$ string                             \\ \cline{2-4}
                & GloVe~\cite{pennington2014glove}    & $n=12, l=300$                                                                     & $300$                              \\ \cline{2-4}
                & MLP1     & $c=324, relu$                                                                     & $324$                              \\ \hline
                \multirow{7}{*}{Visual Embedder} & Input2   & RGB Image & $h \times w \times 3$                   \\ \cline{2-4}
                & Resize   & Uniform Size                                                                      & $224 \times 224 \times 3$                   \\ \cline{2-4}
                & ResNet   & ResNet-$18$                                                                         & $7 \times 7 \times 512$                     \\ \cline{2-4}
                & Conv1    & $c=1024, k=1, bn, relu$               & $7 \times 7 \times 1024$                    \\ \cline{2-4}
                & Dropout  & $rate=0.35$                                                                       & $7 \times 7 \times 1024$                    \\ \cline{2-4}
                & AvgPool  & Global                                                                            & $1 \times 1 \times 1024$                    \\ \cline{2-4}
                & Linear1  & $c=324$                                                                           & $324$                              \\ \hline
                \multirow{25}{*}{OneRestore} & Input3 & RGB Image & $h \times w \times 3$                                                                          \\ \cline{2-4}
                & Input4   & Scene Descriptor                                                                    & $324$                              \\ \cline{2-4}
                & Conv2    & $c=32, k=1$                                                                       & $h \times w \times 32$                      \\ \cline{2-4}
                & SDTB1    & $c=32, head=8$                                                                       & $h \times w \times 32$                      \\ \cline{2-4}
                & Down1    & Maxpool $k=3, s=2$, Conv $c=64,k=1$     & $\frac{h}{2} \times \frac{w}{2} \times 64$  \\ \cline{2-4}
                & SDTB2    & $c=64, head=8$                                                                       & $\frac{h}{2} \times \frac{w}{2} \times 64$  \\ \cline{2-4}
                & Down2    & Maxpool $k=3, s=2$, Conv $c=128,k=1$    & $\frac{h}{4} \times \frac{w}{4} \times 128$ \\ \cline{2-4}
                & SDTB3    & $c=128, head=8$                                                                      & $\frac{h}{4} \times \frac{w}{4} \times 128$ \\ \cline{2-4}
                & Down3    & Maxpool $k=3, s=2$, Conv $c=256,k=1$    & $\frac{h}{8} \times \frac{w}{8} \times 256$ \\ \cline{2-4}
                & SDTB4    & $c=256, head=8$                                                                      & $\frac{h}{8} \times \frac{w}{8} \times 256$ \\ \cline{2-4}
                & SDTB5    & $c=256, head=8$                                                                      & $\frac{h}{8} \times \frac{w}{8} \times 256$ \\ \cline{2-4}
                & SDTB6    & $c=256, head=8$                                                                      & $\frac{h}{8} \times \frac{w}{8} \times 256$ \\ \cline{2-4}
                & Addition1 & Down3 + SDTB6                                                                   & $\frac{h}{8} \times \frac{w}{8} \times 256$ \\ \cline{2-4}
                & SDTB7    & $c=256, head=8$                                                                      & $\frac{h}{8} \times \frac{w}{8} \times 256$ \\ \cline{2-4}
                & Up1      & Bilinear Interpolation, Conv $c=128,k=1$ & $\frac{h}{4} \times \frac{w}{4} \times 128$ \\ \cline{2-4}
                & Addition2 & Up1 + SDTB3                                                                       & $\frac{h}{4} \times \frac{w}{4} \times 128$ \\ \cline{2-4}
                & SDTB8    & $c=128, head=8$                                                                      & $\frac{h}{4} \times \frac{w}{4} \times 128$ \\ \cline{2-4}
                & Up2      & Bilinear Interpolation, Conv $c=64,k=1$  & $\frac{h}{2} \times \frac{w}{2} \times 64$  \\ \cline{2-4}
                & Addition3 & Up2 + SDTB2                                                                       & $\frac{h}{2} \times \frac{w}{2} \times 64$  \\ \cline{2-4}
                & SDTB9    & $c=64, head=8$                                                                       & $\frac{h}{2} \times \frac{w}{2} \times 64$  \\ \cline{2-4}
                & Up3      & Bilinear Interpolation, Conv $c=32,k=1$  & $h \times w \times 32$                      \\ \cline{2-4}
                & Addition4 & Up3 + SDTB1                                                                       & $h \times w \times 32$                      \\ \cline{2-4}
                & SDTB10    & $c=32, head=8$                                                                       & $h \times w \times 32$                      \\ \cline{2-4}
                & Conv3    & $c=3, k=1$                                                                        & $h \times w \times 3$                       \\ \cline{2-4}
                & Addition5 & Conv3 + Input3                                                                    & $h \times w \times 3$                       \\ \hline \hline      
        \end{tabular}
        \vspace{-3mm}
    \end{table}
    \begin{algorithm}[t]
    \caption{Text/Visual Embedder Training}
    \label{alg:embedder_train}
        \begin{algorithmic}[1]
            \renewcommand{\algorithmicrequire}{\textbf{Input:}}
    	\Require
    	Visual image and scene description text label pairs $(I_v, y)$, all texts $\mathcal{S}_t$.
    	\Repeat
     	\State $e_t=Emb_{t}(\mathcal{S}_t)$ \Comment{Get text embeddings.}
     	\State $e_v=Emb_{v}(I_v)$ \Comment{Get visual embeddings.}
            \State Perform the gradient descent step by \newline
     \centerline{$\mathcal{L}_{cross} = -\frac{1}{N_v} \sum_{i=1}^{N_v}\sum_{j=1}^{N_t}y_{ij}\log{(S(e_{v_i},e_{t_j}))}$.} \Comment{$N_v$ and $N_t$ are the numbers of visual and text embeddings, respectively. $y_{i}$ is the truth label of $i$-th sample. When the truth label is $j$, $y_{ij}=1$, otherwise $y_{ij}=0$.}
            \Until converged
            \State \textbf{return} $Emb_{t}$, $Emb_{v}$
        \end{algorithmic}
    \end{algorithm}
    \begin{algorithm}[t]
    \caption{OneRestore Model Training}
    \label{alg:OneRestore_train}
        \begin{algorithmic}[1]
            \renewcommand{\algorithmicrequire}{\textbf{Input:}}
    	\Require
    	Clear, input degraded, and other degraded image pairs $(J, I, \{I_o\})$, scene description text of input degraded image $\mathcal{S}_t' \in \mathcal{S}_t$, text embedder $Emb_{t}$.
            \Repeat
            \State $e_t'=Emb_{t}(\mathcal{S}_t')$ \Comment{Get the scene descriptor with the frozen text embedder.}
            \State $\hat{J}=OneRestore(I,e_t')$ \Comment{Get the restored image.}
            \State Perform the gradient descent step by \newline \centerline{$\mathcal{L} = \alpha_1 \mathcal{L}^s_{1}(J,\hat{J})+ \alpha_2 \mathcal{L}_\text{M}(J,\hat{J})+ \alpha_3\mathcal{L}_\text{c}(J,\hat{J},I,\{I_o\})$.}
            \Until converged
            \State \textbf{return} $OneRestore$
        \end{algorithmic}
    \end{algorithm}
    \begin{algorithm}[t]
    \caption{Model Inference}
    \label{alg:OneRestore_test}
        \begin{algorithmic}[1]
            \renewcommand{\algorithmicrequire}{\textbf{Input:}}
    	\Require
    	Input degraded image $I$, scene description text of input degraded image $\mathcal{S}_t' \in \mathcal{S}_t$ (non-essential).
            \If{exist $\mathcal{S}_t'$} \Comment{Manual restoration.}
            \State $e_t'=Emb_{t}(\mathcal{S}_t')$
            \Else \Comment{Automatic restoration.}
            \State $e_v'=Emb_{v}(I)$
            \State Calculate the cosine similarity between $e_v'$ and each text embedding by \newline \centerline{$cos(e_v', e_t) = \gamma \cdot \frac{e_v' \cdot e^\top _t}{\left \|  e_v'\right \|\left \|  e_t\right \| }$.}
            \State Select the text embedding with the highest similarity as the estimated scene descriptor $e_t'$.
            \EndIf
            \State $\hat{J}=OneRestore(I,e_t')$
            \State \textbf{return} $\hat{J}$
        \end{algorithmic}
    \end{algorithm}
    \vspace{-3mm}
    \begin{figure}[h]
        \centering
        \includegraphics[width=0.95\linewidth]{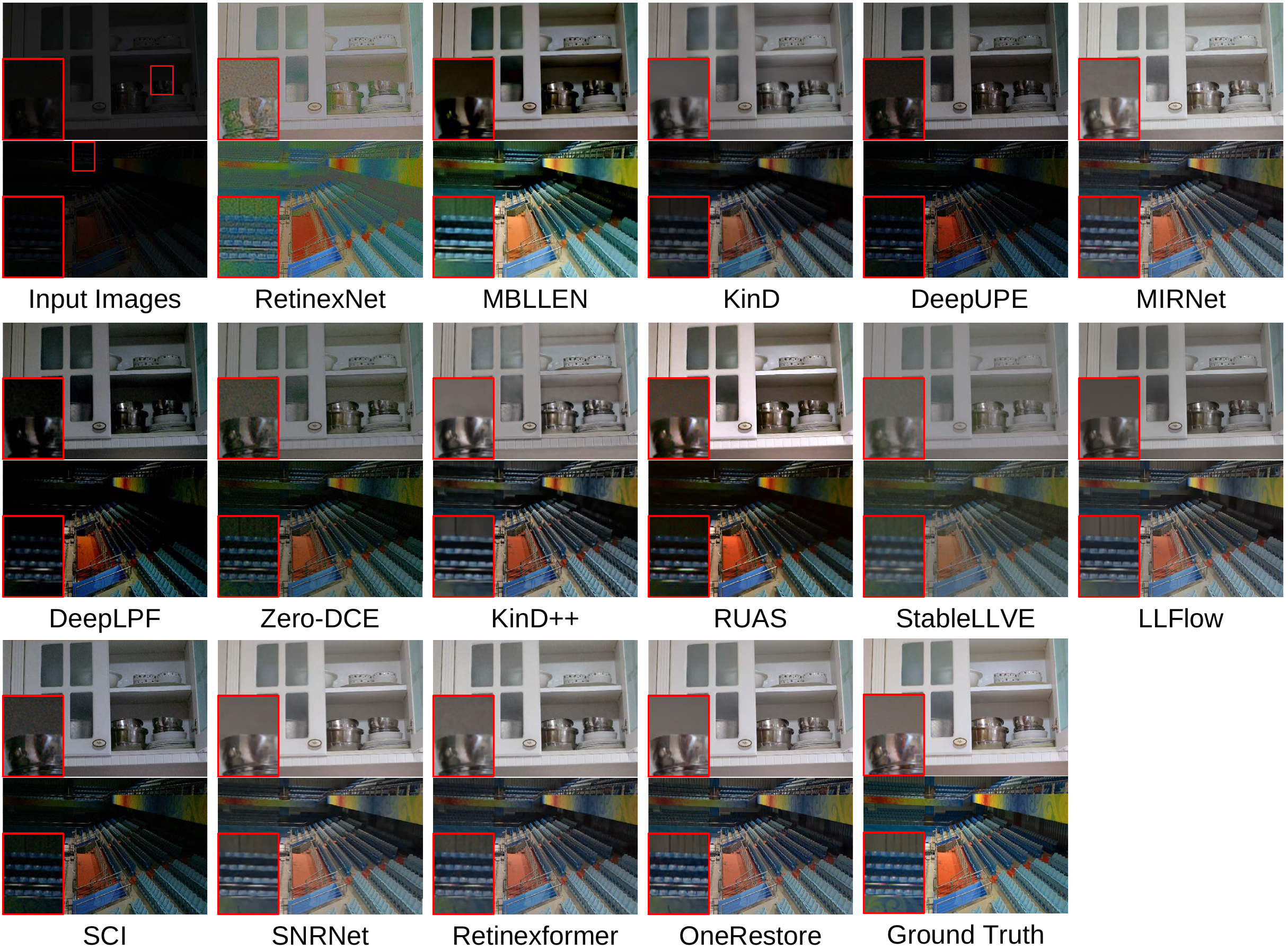}
        \vspace{-2mm}
        \caption{Comparison of low-light enhancement on LOL dataset~\cite{Chen2018Retinex}.}
        \label{fig:lol}
        \vspace{-2mm}
    \end{figure}
    \begin{figure}[ht]
        \centering
        \includegraphics[width=1\linewidth]{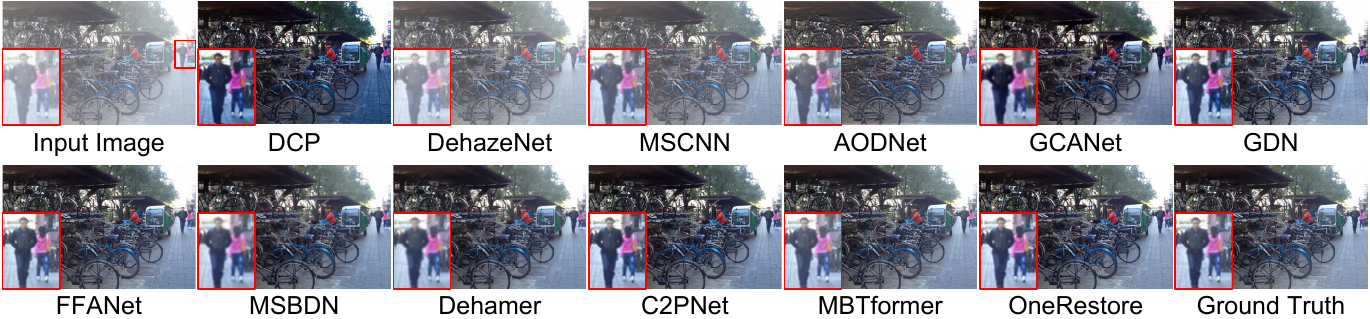}
        \vspace{-3mm}
        \caption{Comparison of image dehazing on RESIDE dataset~\cite{li2018benchmarking}.}
        \label{fig:reside}
        \vspace{-3mm}
    \end{figure}
\subsection{Detailed Architecture}
    All network configurations and output sizes of each layer are displayed in Table~\ref{tb:config}. Specifically, the MLP layer used in the text embedder consists of a linear operator and a ReLU activation function. The visual embedder introduces a dropout layer to prevent overfitting. Our OneRestore employs three max-pooling operations to downsample images and utilizes the Scene Descriptor-guided Transformer Block (SDTB) to extract multi-scale image features. Finally, bilinear interpolation upsampling and skip connections are employed to fuse features of different scales and levels.
\section{Model Training and Inference}
    The model training includes two steps: text/visual embedder training and OneRestore model training. Meanwhile, the model inference involves a manual mode based on text embedding and an automatic mode controlled by visual attributes.
\subsection{Step1: Text/Visual Embedder Training.} 
    The pseudo-code is shown in Alg. \ref{alg:embedder_train}. Initially, we set RGB image and scene description text label pairs as $(I_v, y)$ and all texts as $\mathcal{S}_t$, respectively. Subsequently, the text embedder $Emb_{t}$ is employed to generate 12 scene description embeddings $e_t$ (line 2 in Alg. \ref{alg:embedder_train}), and the visual embedder $Emb_{v}$ is utilized to produce visual embeddings $e_v$ from $I_v$ (line 3 in Alg. \ref{alg:embedder_train}). Finally, we calculate the cosine similarity between visual embeddings and text embeddings and perform gradient descent via the cross-entropy loss (line 4 in Alg. \ref{alg:embedder_train}). 
\subsection{Step2: OneRestore Model Training.} 
    The pseudo-code is shown in Alg. \ref{alg:OneRestore_train}. We consider the clear, input degraded, and other degraded image pairs $(J, I, \{I_o\})$, scene description text of the input degraded image $\mathcal{S}_t' \in \mathcal{S}_t$, and pre-trained text embedder $Emb_{t}$ as inputs. $Emb_{t}$ is first used to generate scene descriptors $e_t'$ (line 2 in Alg. \ref{alg:OneRestore_train}). Input degraded images $I$ and scene descriptors $e_t'$ are jointly fed into our OneRestore to generate restored results (line 3 in Alg. \ref{alg:OneRestore_train}). For the model converges, the total loss $\mathcal{L}$ is employed for gradient descent (line 3 in Alg. \ref{alg:OneRestore_train}). 
    \begin{figure}[htpb]
        \centering
        \includegraphics[width=1\linewidth]{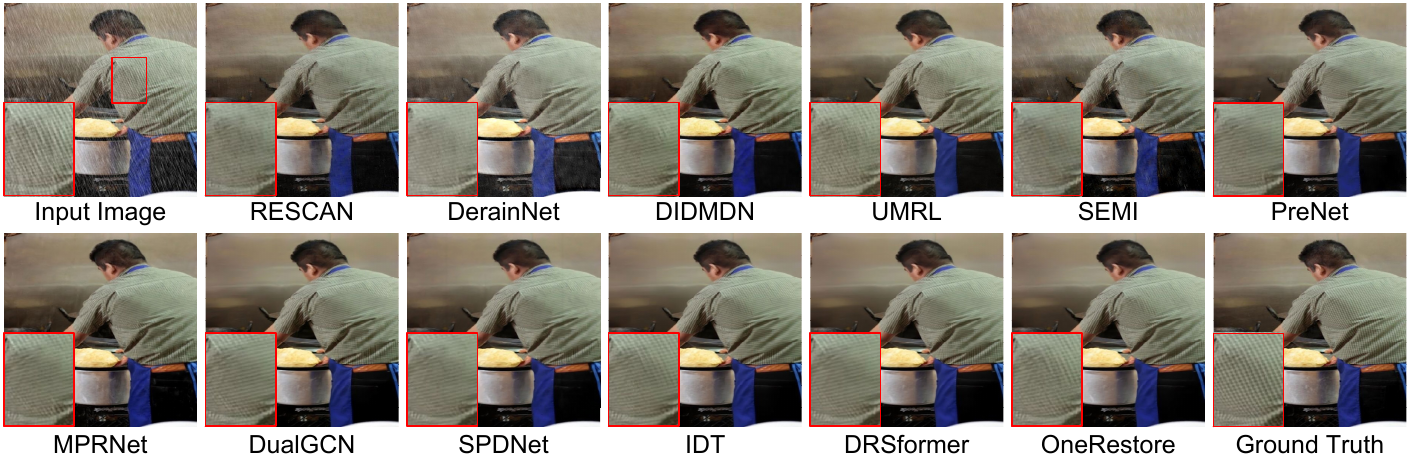}
        \caption{Comparison of image dehazing on Rain1200 dataset~\cite{zhang2018density}.}
        \label{fig:rain1200}
    \end{figure}
    \begin{figure}[htpb]
        \centering
        \includegraphics[width=1\linewidth]{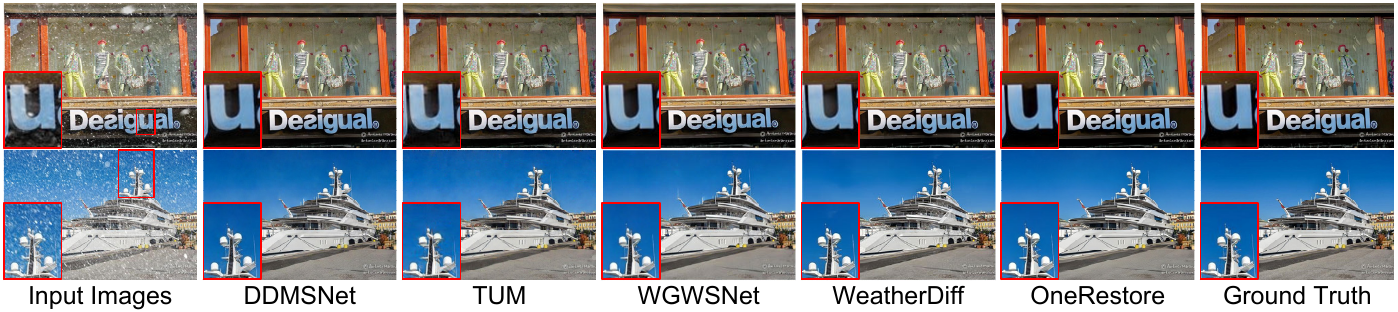}
        \vspace{-6mm}
        \caption{Comparison of image desnowing on Snow100k-L dataset~\cite{liu2018desnownet}.}
        \vspace{-4mm}
        \label{fig:snow100k}
    \end{figure}
    \begin{figure}[htpb]
        \centering
        \includegraphics[width=1\linewidth]{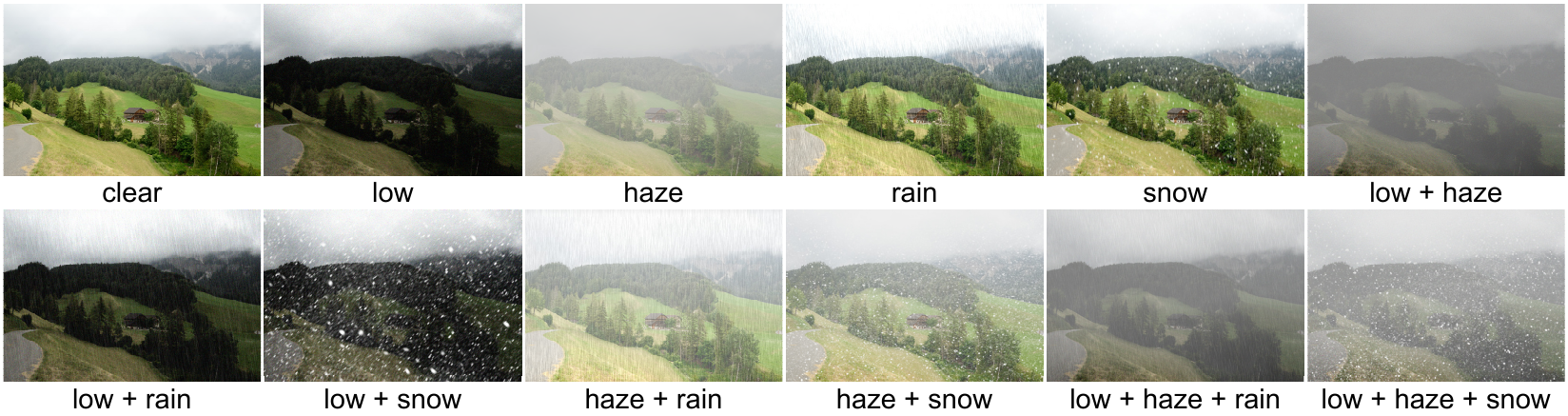}
        \caption{Visualization of one test sample on CDD-11 dataset.}
        \label{fig:syn}
    \end{figure}
\begin{table}[ht]
    \centering
    \caption{Comparison of quantitative results on four benchmarks. {\color{red} Red}, {\color{green} green}, and {\color{blue} blue} indicate the best, second-best, and third-best results, respectively.}
    \vspace{-3mm}
    \resizebox{\textwidth}{!}{%
    \begin{tabular}{lccc|lccc}
    \hline\hline
        Methods & Venue \& Year & PSNR $\uparrow$ & SSIM $\uparrow$& Methods & Venue \& Year & PSNR $\uparrow$& SSIM $\uparrow$\\ \hline
        Input & & 7.77 & 0.1914 & Input & & 15.92 & 0.8139 \\ \hline
        RetinexNet~\cite{Chen2018Retinex} & BMVC2018 & 17.12 & 0.5920 & DCP~\cite{he2010single} & TPAMI2010 & 14.67 & 0.7825 \\
        MBLLEN~\cite{lv2018mbllen} & BMVC2018 & 17.86 & 0.7247 & DehazeNet~\cite{cai2016dehazenet} & TIP2016 & 20.95 & 0.8841\\
        KinD~\cite{zhang2019kindling} & ACMMM2019 & 17.71 & 0.7734 & MSCNN~\cite{ren2016single} & ECCV2016 & 20.00 & 0.8779 \\
        MIRNet~\cite{zamir2020learning} & ECCV2020 & 24.10 & {\color{blue}0.8421} & AODNet~\cite{li2017aod} & ICCV2017 & 20.36 & 0.8945 \\
        Zero-DCE~\cite{guo2020zero} & CVPR2020 & 14.86 & 0.5624 & GCANet~\cite{chen2019gated} & WACV2019 & 22.60 & 0.8985 \\
        KinD++~\cite{zhang2021beyond} & IJCV2021 & 17.75 & 0.7581 & GDN~\cite{liu2019griddehazenet} & ICCV2019 & 30.77 & 0.9808 \\
        RUAS~\cite{liu2021retinex} & CVPR2021 & 16.40 & 0.5034 & FFANet~\cite{qin2020ffa} & AAAI2020 & 32.13 & 0.9792\\
        StableLLVE~\cite{zhang2021learning} & CVPR2021 & 17.36 & 0.7373 & MSBDN~\cite{dong2020multi} & CVPR2020 & 30.23 & 0.9458\\
        LLFlow~\cite{wang2022low} & AAAI2022 & 19.34 & 0.8388 & DeHamer~\cite{guo2022image} & CVPR2022 & 30.70 & 0.9457\\
        SNRNet~\cite{xu2022snr} & CVPR2022 & {\color{green}24.61} & 0.8401 & C2PNet~\cite{zheng2023curricular} & CVPR2023 & {\color{blue} 34.05} & {\color{green} 0.9857} \\
        Retinexformer~\cite{cai2023retinexformer} & ICCV2023 & {\color{red}25.15} & {\color{green}0.8434} & MB-TFormer~\cite{qiu2023mb} & ICCV2023 & {\color{red} 37.94} & {\color{red} 0.9899} \\ \hline
        OneRestore & & {\color{blue}24.25} & {\color{red}0.8564} & OneRestore & & {\color{green} 35.58} & {\color{blue} 0.9814} \\ \hline\hline
        \multicolumn{4}{c}{\textbf{(a) Enhancement results on the LOL dataset~\cite{Chen2018Retinex}.}\vspace{1mm}} & \multicolumn{4}{c}{\textbf{(b) Dehazing results on the RESIDE-OTS dataset~\cite{li2018benchmarking}.}\vspace{1mm}} \\ \hline\hline
        Methods & Venue \& Year & PSNR $\uparrow$& SSIM $\uparrow$& Methods & Venue \& Year & PSNR $\uparrow$& SSIM $\uparrow$\\ \hline
        Input & & 22.16 & 0.6869 &Input & & 18.68 & 0.6470 \\ \hline
        RESCAN~\cite{li2018recurrent} & ECCV2016 & 28.83 & 0.8430 & DerainNet~\cite{fu2017clearing} & TIP2017 & 19.18 & 0.7495 \\
        DerainNet~\cite{fu2017clearing} & TIP2017 & 21.93 & 0.7814 & DehazeNet~\cite{cai2016dehazenet} & TIP2016 & 22.62 & 0.7975  \\
        DID-MDN~\cite{zhang2018density} & CVPR2018 & 27.99 & 0.8627 & DeepLab~\cite{chen2017deeplab} & TPAMI2017 & 21.29 & 0.7747 \\
        UMRL~\cite{yasarla2019uncertainty} & CVPR2019 & 28.62 & 0.8706 & RESCAN~\cite{li2018recurrent} & ECCV2018 & 26.08 & 0.8108 \\
        SEMI~\cite{wei2019semi} & CVPR2019 & 24.39 & 0.7622 & SPANet~\cite{wang2019spatial} & CVPR2019 & 23.70 & 0.7930 \\
        PreNet~\cite{ren2019progressive} & CVPR2019 & 29.79 & 0.8811 & AIO~\cite{li2020all} & CVPR2020 & 28.33 & 0.8820 \\
        MPRNet~\cite{zamir2021multi} & CVPR2021 & 31.32 & 0.8907 & DDMSNet~\cite{zhang2021deep} & TIP2021 & 28.85 & 0.8772 \\
        DualGCN~\cite{fu2021rain} & AAAI2021 & 32.09 & 0.9181 & TransWeather~\cite{valanarasu2022transweather} & CVPR2022 & {\color{blue}29.31} & {\color{blue}0.8879} \\
        SPDNet~\cite{yi2021structure} & ICCV2021 & 32.84 & 0.9138 & TUM~\cite{chen2022learning} & CVPR2022 & 26.90 & 0.8321 \\
        IDT~\cite{xiao2022image} & TPAMI2022 & {\color{green}33.13} & {\color{green}0.9238} & WGWSNet~\cite{zhu2023learning} & CVPR2023 & 28.94 & 0.8758 \\
        DRSformer~\cite{chen2023learning} & CVPR2023 & {\color{red}33.59} & {\color{red}0.9274} & WeatherDiff~\cite{ozdenizci2023restoring} & TPAMI2023 & {\color{green}30.09} & {\color{red}0.9041} \\ \hline
        OneRestore & & {\color{blue}32.89} & {\color{blue}0.9182} & OneRestore & & {\color{red}30.24} & {\color{green}0.8947} \\ \hline \hline
        \multicolumn{4}{c}{\textbf{(c) Deraining results on the Rain1200 dataset~\cite{zhang2018density}.}\vspace{1mm}} & \multicolumn{4}{c}{\textbf{(d) Desnowing results on the Snow100k dataset~\cite{liu2018desnownet}.}} \\
    \end{tabular}%
    }\vspace{-6mm}
    \label{tb:benchmarks}
\end{table}
\setlength{\tabcolsep}{8pt}
\begin{table}[ht]
    \centering
    \caption{Comparison of quantitative results on four real benchmarks. The best results are in \textbf{bold}, and the second-best are with \underline{underline}.}
    \vspace{-3mm}
    \resizebox{\textwidth}{!}{%
    \begin{tabular}{lccc|lccc@{}}
    \hline \hline
        Methods                      & Venue \& Year & NIQE $\downarrow$ & PIQE $\downarrow$ & Methods                           & Venue \& Year & NIQE $\downarrow$& PIQE $\downarrow$ \\ \hline
RUAS~\cite{liu2021retinex}   & IJCV2021      & 7.77 & 19.64 & MSBDN~\cite{dong2020multi}        & CVPR2020      & \underline{4.77} & 26.14 \\
SCI~\cite{ma2022toward}      & CVPR2021      & \underline{3.97} & \textbf{16.35} & DeHamer~\cite{guo2022image}       & CVPR2022      & 5.34 & 31.93 \\
SNRNet~\cite{xu2022snr}      & CVPR2022      & 4.49 & 20.71 & C2PNet~\cite{zheng2023curricular} & CVPR2023      & 5.03 & \underline{25.09} \\ 
OneRestore                   &               & \textbf{3.93} & \underline{16.69} & OneRestore                        &               & \textbf{4.58} & \textbf{23.91} \\ \hline \hline
\multicolumn{4}{c}{\textbf{(a) Enahncement results on the NPE dataset~\cite{wang2013naturalness}.}\vspace{1mm}}          & \multicolumn{4}{c}{\textbf{(b) Dehazing results on the RTTS dataset~\cite{li2018benchmarking}.}\vspace{1mm}}               \\
\hline \hline
 Methods                      & Venue \& Year & NIQE $\downarrow$& PIQE $\downarrow$ & Methods                           & Venue \& Year & NIQE $\downarrow$& PIQE $\downarrow$ \\ \hline
MPRNet~\cite{zamir2021multi} & CVPR2021      & 3.55 & 20.78 & DRT~\cite{liang2022drt}       & CVPRW2022      & 3.93 & 12.00  \\
DualGCN~\cite{fu2021rain}    & AAAI2021      & \textbf{3.27} & \underline{15.15} & TUM~\cite{chen2022learning}       & CVPR2022      & \underline{3.13} & \underline{9.14} \\
MFDNet~\cite{wang2023multi}  & TIP2023       & 3.37 & 18.86 &  UMWT~\cite{kulkarni2022unified} & ECCVW2022 & 3.34 & 12.29 \\
OneRestore                   &               & \underline{3.32} & \textbf{14.46} &                                   OneRestore                        &               & \textbf{3.00} & \textbf{9.12}    \\ \hline \hline
\multicolumn{4}{c}{\textbf{(c) Deraining results on the RS dataset~\cite{yang2017deep}.}}                  & \multicolumn{4}{c}{\textbf{(d) Desnowing results on the Snow100k-R dataset~\cite{liu2018desnownet}.}}   \\ 
    \end{tabular}}
    \label{tb:real}
\end{table}
\begin{figure}[htpb]
    \centering
    \includegraphics[width=1\linewidth]{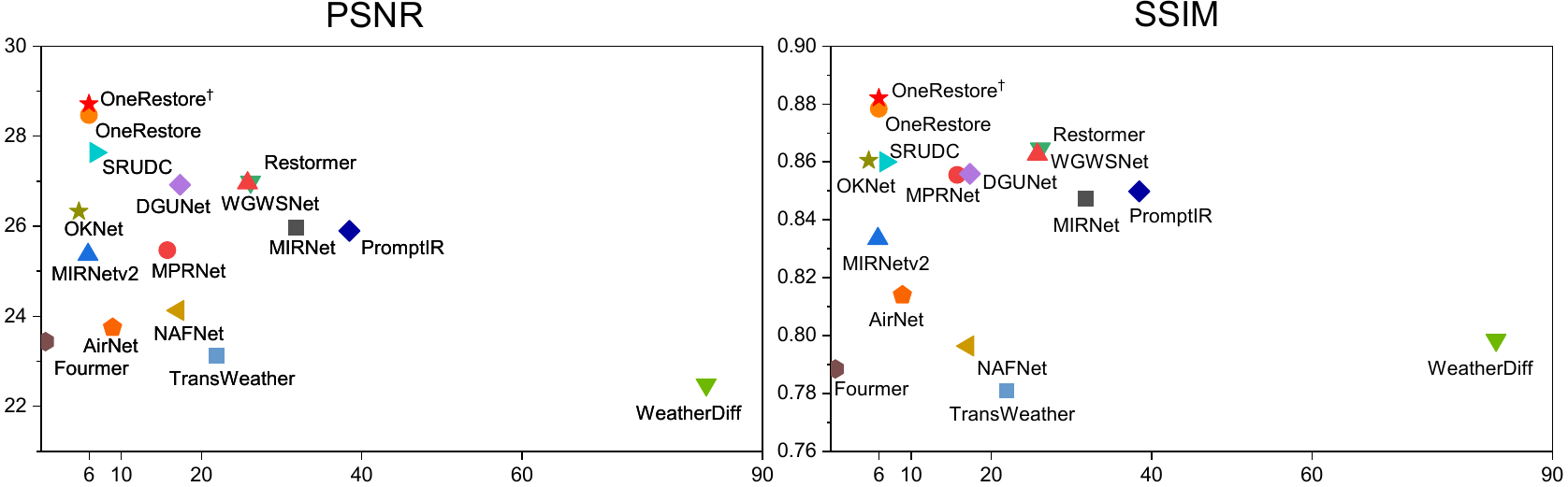}\\
    \vspace{-1mm}
    \centering
    \caption{Comparison of quantitative results and parameter quantities on CDD-11 dataset. OneRestore$^\dagger$ indicates the manual mode, while OneRestore denotes the automatic mode.}
    \vspace{-4mm}
    \label{fig:metric2}
\end{figure}
\subsection{Model Inference.} 
    As shown in Alg. \ref{alg:OneRestore_test}, our model inference has two modes: manual and automatic. The necessary input is the degraded image $I$, and the optional input is the scene description text $\mathcal{S}_t' \in \mathcal{S}_t$. When the manual mode is selected, scene descriptors $e_t'$ will be generated directly from $\mathcal{S}_t'$ using $Emb_{t}$ (line 2 in Alg. \ref{alg:OneRestore_test}). When using the automatic mode, $Emb_{v}$ will first generate visual embedding $e_v'$ (line 4 in Alg. \ref{alg:OneRestore_test}). Then, the distance between the visual embedding $e_v'$ and all text embeddings $e_t$ is calculated by cosine similarity, and the top-1 text embedding is selected as the OneRestore input (line 5-6 in Alg. \ref{alg:OneRestore_test}). The restored result will be generated by processing $I$ and $e_t'$ through our OneRestore model (line 8 in Alg. \ref{alg:OneRestore_test}).
    \begin{figure*}[p]
        \centering
        \includegraphics[width=1\linewidth]{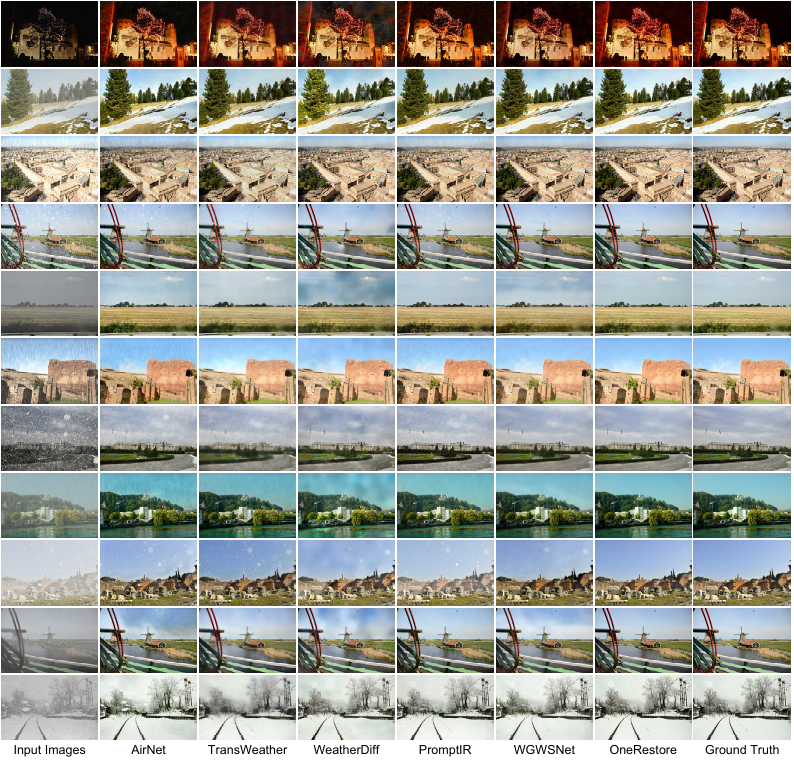}\\
        \caption{Comparison of image restoration on \textit{low}, \textit{haze}, \textit{rain}, \textit{snow}, \textit{low+haze}, \textit{low+rain}, \textit{low+snow}, \textit{haze+rain}, \textit{haze+snow}, \textit{low+haze+rain}, and \textit{low+haze+snow} synthetic samples.}
        \label{fig:enhance_syn1}
    \end{figure*}
    \begin{figure*}[p]
        \centering
        \includegraphics[width=1\linewidth]{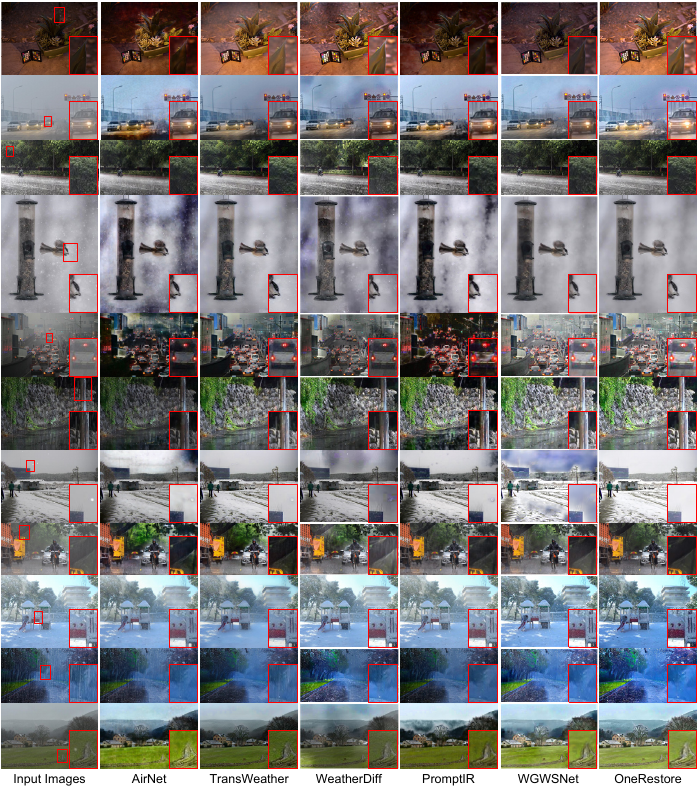}\\
        \caption{Comparison of image restoration on \textit{low}, \textit{haze}, \textit{rain}, \textit{snow}, \textit{low+haze}, \textit{low+rain}, \textit{low+snow}, \textit{haze+rain}, \textit{haze+snow}, \textit{low+haze+rain}, and \textit{low+haze+snow} samples in real-world scenarios.}
        \label{fig:enhance_real1}
    \end{figure*}
\begin{figure}[t]
    \centering
    \begin{minipage}[t]{0.49\textwidth}
    \centering
    \includegraphics[width=1\linewidth]{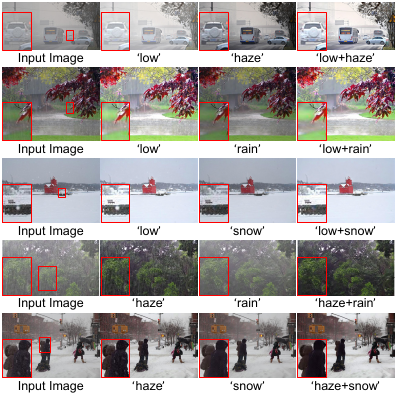}
    \end{minipage}
    \begin{minipage}[t]{0.49\textwidth}
    \centering
    \includegraphics[width=1\linewidth]{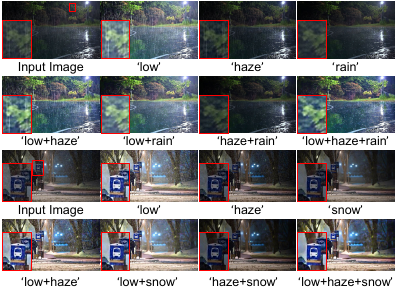}
    \end{minipage}
    {\begin{scriptsize}
    \begin{tabular}{cc}
        \makebox[0.45\textwidth][c]{(a) Double degeneration coexistence examples} & \makebox[0.45\textwidth][c]{(b) Triple degenerate coexistence examples}
    \end{tabular} \end{scriptsize}}\\
    \vspace{-3mm}
    \caption{Comparison of image restoration on real-world scenarios by using different text descriptors.}
    \vspace{-5mm}
    \label{fig:control3}
\end{figure}
\begin{figure*}[htpb]
    \centering
    \includegraphics[width=1\linewidth]{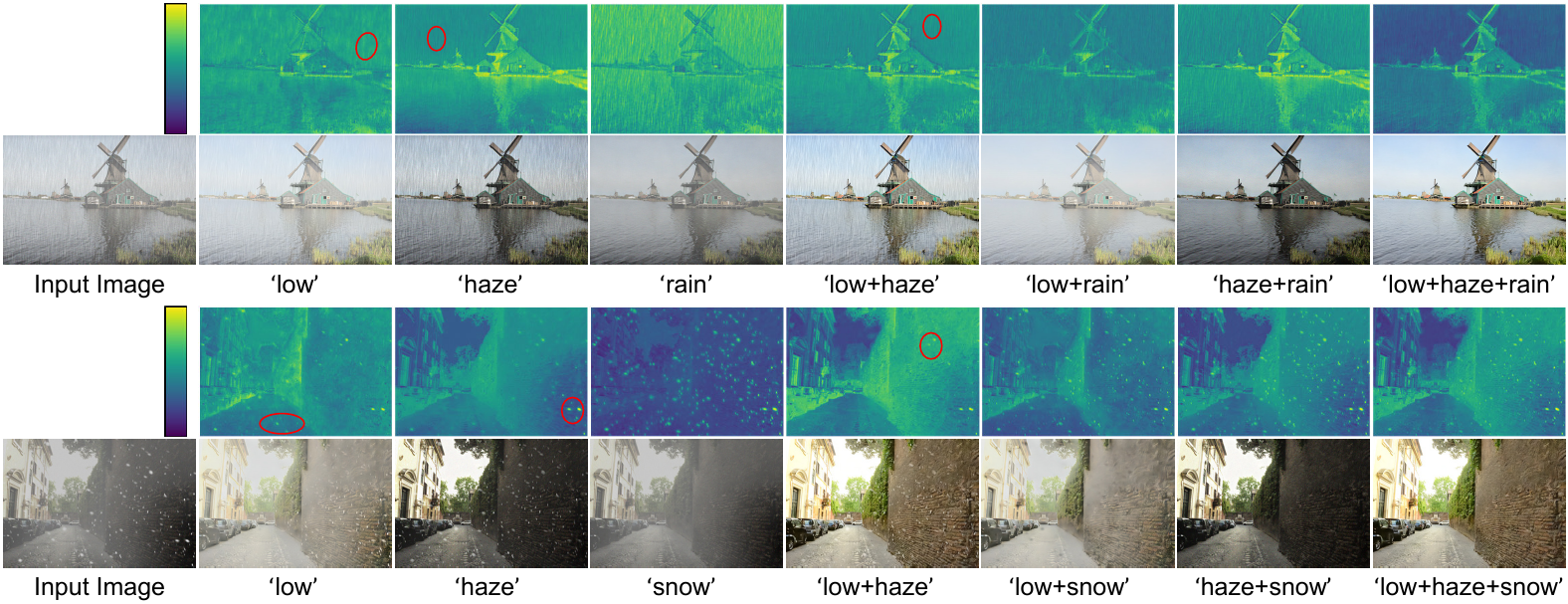}\\
    \vspace{-1mm}
    \caption{Comparison of the SDTB10 output feature maps and restoration results based on different texts on \textit{low+haze+rain} and \textit{low+haze+snow} synthetic samples. Due to the intricate interplay of different degradation factors, our model tends to slightly eliminate features not explicitly mentioned in the description text. As indicated in the red oval regions, this phenomenon is often observed in the context of rain and snow degradation. Nevertheless, our model already exhibits impressive controllability.}
    \vspace{-2mm}
    \label{fig:control4}
\end{figure*}
\section{Experiments Results}
    In this section, we conduct a comparative analysis of our OneRestore against state-of-the-art (SOTA) models on benchmark datasets of related tasks, aiming to validate OneRestore's generalization ability. Subsequently, extensive experiments of image restoration, encompassing both synthetic and real-world scenarios, are presented to further illustrate the method's efficacy. Lastly, an analysis focusing on the model controllability is performed.
\subsection{Comparison on Classic Benchmarks}
    We train and test our OneRestore on classic benchmark datasets for low-light enhancement (LOL~\cite{Chen2018Retinex}), dehazing (RESIDE-OTS ~\cite{li2018benchmarking}), draining (Rain1200~\cite{zhang2018density}), and desnowing (Snow100k~\cite{liu2018desnownet}), and conduct a comprehensive comparison of our method with SOTA methods in each task. The quantitative results and visual performance of each method are reported in Table~\ref{tb:benchmarks} and Figs.~\ref{fig:lol}-\ref{fig:snow100k}, respectively. Given that our OneRestore is designed to address composite degradations, it is inevitable that it cannot surpass the current optimal SOTA methods for all benchmarks. Nonetheless, the comparable performance of the recovery results unequivocally demonstrates the model's robust feature extraction capabilities. It is noteworthy that our method boasts a relatively smaller number of parameters (only 5.98M) and faster inference speed (0.0115s for processing a 1080$\times$720 image on our PC), further reinforcing its practical significance.
\subsection{More Results on CDD-11 Dataset}
    We conduct experiments on the proposed Composite Degradation Dataset (CDD-11), where an example of a clear image and its corresponding 11 types of degraded samples are shown in Fig.~\ref{fig:syn}. Quantitative comparisons of all methods on our CDD-11 dataset are shown in Fig.~\ref{fig:metric2}, demonstrating that our OneRestore can achieve a balance between quantitative results and parameter quantities.

    Furthermore, Fig.~\ref{fig:enhance_syn1} displays more restoration cases on the 11 types of degraded images from the CDD-11 dataset. It is clear that current SOTA methods are limited in their ability to handle all types of image degradation and can produce unstable results. In contrast, our method is designed to be versatile and can effectively handle a wide range of degradation scenarios. By incorporating scene descriptors to identify degradation situations, our method can create high-quality images with rich details.
\subsection{More Results on Real-World Dataset}
    To assess the robustness of our model, which was trained using the CDD-11 dataset, against complex degradation scenarios, we conducted extensive real-world image restoration experiments. Our quantitative analysis on four distinct real-world benchmarks utilized two no-reference quality assessment metrics: the Natural Image Quality Evaluator (NIQE) ~\cite{mittal2012making} and the Perception-based Image Quality Evaluator (PIQE) ~\cite{venkatanath2015blind}. It was noted that these benchmarks were specifically chosen to represent the four degradation challenges we aimed to address: the Naturalness Preserved Enhancement dataset (NPE)~\cite{wang2013naturalness} for low-light enhancement, the RESIDE Real-world Task-driven Testing Set (RTTS)~\cite{li2018benchmarking} for image dehazing, Yang's dataset (RS)~\cite{yang2017deep} for image deraining, and the real Snow100k dataset (Snow100k-R)~\cite{liu2018desnownet} for image desnowing. As shown in Table \ref{tb:real}, our quantitative assessment comparison with SOTA methods substantiates the exceptional performance of our model in restoring these degradation datasets. Moreover, we show additional real-world image restoration cases in Fig.~\ref{fig:enhance_real1}. It is evident that models trained on the CDD-11 dataset exhibit remarkable efficacy in real-world scenarios, thereby validating the proposed synthesis strategy as an effective simulation tool for real composite degradation scenarios. Notably, both WGWSNet and our method, which introduces degraded descriptions, demonstrate superior visual effects. Our OneRestore establishes a more stringent lower-bound constraint by applying the composite degradation restoration loss, enabling its restoration results to closely approximate clear scenarios.
    
\subsection{More Results of Different Scene Descriptors}
    To verify the controllability of our model, we conduct a comparison experiment on seven distinct types of images employing varying text embeddings. The resulting visual outcomes are shown in Fig.~\ref{fig:control3}. By introducing scene description texts, we can direct the model's attention to different types of degradation factors, thereby achieving controllable restoration. Furthermore, the input of multiple words in conjunction enables our OneRestore to achieve better visual performance in complex composite degradation scenarios. Fig. \ref{fig:control4} further illustrates the impact of using different scene description texts on the feature maps output by SDTB10 and restoration results under \textit{low+haze+rain} and \textit{low+haze+snow} synthetic samples. The introduction of scene description texts can serve as prior information, enabling the model to establish the distinct representation space for correspondence degradation. This characteristic positions our OneRestore as a superior solution in addressing complex, composite degradations.
    
\section{Limitation and Future Work}
    Extensive experiments have been conducted to verify the proposed method's performance on both synthetic and real composite degradation scenarios. Impressive results have demonstrated that the proposed unified imaging model effectively simulates real-world complex degradation. Moreover, the introduction of degraded scene descriptors and the proposed composite degradation restoration loss can aid the model in adjusting to complex situations. However, our approach still has failure recovery cases, as illustrated in Fig.~\ref{fig:failure}. Although some degradation factors are well suppressed or eliminated, color abnormalities and distortions still exist in the restoration results. Based on the observation of failure results, we elaborate on the limitations and future research directions of this work in more detail, which can be introduced as follows:
\begin{figure}[t]
    \centering
    \includegraphics[width=0.7\linewidth]{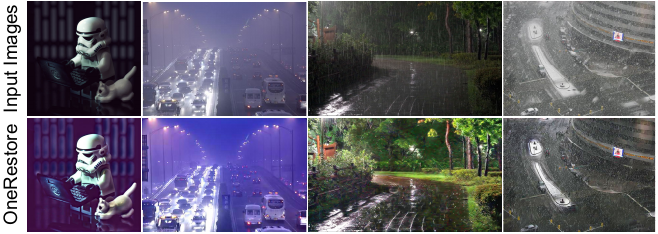}\\
    \vspace{-2mm}
    \caption{Failure cases of image restoration.}
    \vspace{-5mm}
    \label{fig:failure}
\end{figure}
    \begin{itemize}
        \item Disparities between synthetic and real-world data may constrain the image restoration ability of our approach in some real degradation scenarios, such as high-density corruption scenarios.
        \item More degradation types are not considered in this work, such as raindrops, moiré, shadows, etc. However, the proposed universal framework provides a solid solution, making it possible to cover a wider range of degradation types by incorporating additional scene descriptors for directional restoration.
        \item While the proposed approach provides restoration control over the degradation type, achieving restoration intensity control for each type remains a direction for future research.
        \item Ensuring model robustness while reducing computational overhead is also one of our future considerations.
    \end{itemize}

\end{document}